\documentclass{article}

\usepackage[final,nonatbib]{neurips_2022}

\usepackage[utf8]{inputenc} % allow utf-8 input
\usepackage[T1]{fontenc}    % use 8-bit T1 fonts
\usepackage{hyperref}
\hypersetup{
	colorlinks   = true, %Colours links instead of ugly boxes
	urlcolor     = blue, %Colour for external hyperlinks
	linkcolor    = red, %Colour of internal links
	citecolor   = blue %Colour of citations
}

\usepackage{url}            % simple URL typesetting
\usepackage{booktabs}       % professional-quality tables
\usepackage{amsfonts}       % blackboard math symbols
\usepackage{nicefrac}       % compact symbols for 1/2, etc.
\usepackage{microtype}      % microtypography
\usepackage{xcolor}         % colors

\usepackage{times}
\usepackage[small]{caption}
\usepackage{graphicx}
\usepackage{amsmath,amssymb}
\usepackage{amsthm}
\usepackage{booktabs}
\usepackage{algorithm}
\usepackage{algorithmic}
\usepackage{bm,enumerate}

\usepackage{subcaption}

\usepackage{array}
\usepackage{adjustbox}
\usepackage{makecell}

\newcolumntype{H}{>{\setbox0=\hbox\bgroup}c<{\egroup}@{}}
\newcolumntype{C}{>{\centering\arraybackslash}m{1cm}}

\newcommand{\indep}{\perp \!\!\! \perp}

\newtheorem{theorem}{Theorem}

\title{Multi-Instance Causal Representation Learning for Instance Label Prediction and Out-of-Distribution Generalization}

\author{%
 Weijia Zhang$^{1\thanks{Corresponding author}}$, \space Xuanhui Zhang$^2$,  \space Han-Wen Deng$^1$,  \space Min-Ling Zhang$^1$
 \\
  $^1$School of Computer Science and Engineering, Southeast University, Nanjing 210096, China\\
  $^2$ School of Information Management, Nanjing University, Nanjing 210023, China\\
  zhangwj@seu.edu.cn, zhangxhdo@163.com,  \{denghw,zhangml\}@seu.edu.cn\\
}

\begin{document}

\maketitle

\begin{abstract}
Multi-instance learning (MIL) deals with objects represented as bags of instances and can predict instance labels from bag-level supervision. 
However, significant performance gaps exist between instance-level MIL algorithms and supervised learners since the instance labels are unavailable in MIL. 
Most existing MIL algorithms tackle the problem by treating multi-instance bags as harmful ambiguities and predicting instance labels by reducing the supervision inexactness.
This work studies MIL from a new perspective by considering bags as auxiliary information, and utilize it to identify instance-level causal representations from bag-level weak supervision. 
We propose the CausalMIL algorithm, which not only excels at instance label prediction but also provides robustness to distribution change by synergistically integrating MIL with identifiable variational autoencoder. 
Our approach is based on a practical and general assumption: 
the prior distribution over the instance latent representations belongs to the non-factorized exponential family conditioning on the multi-instance bags.
Experiments on synthetic and real-world datasets demonstrate that our approach significantly outperforms various baselines on instance label prediction and out-of-distribution generalization tasks.
\end{abstract}

\section{Introduction}
Supervised learning has achieved great success in many applications. However, an important limitation of existing supervised learning algorithms is that they model complex objects as a single feature vector and thus cannot make fine-grained predictions without the corresponding level of supervision, e.g., localizing the region of interest within an image from coarse-grained image-level supervision.
Unfortunately, acquiring fine-grained labels is often not only a tedious task but also prohibitively expensive, especially for applications that require a high level of domain-specific expertise such as drug activity prediction \cite{Dietterich1997} and medical image classification \cite{Lu2021}. 

Multi-Instance Learning (MIL) \cite{Dietterich1997} is a weakly supervised learning paradigm originally proposed for drug activity prediction, where the task is to predict whether a molecule is suitable for binding to a target receptor. 
Since a molecule can take many low-energy conformations and its suitability for making drugs depends on some specific but unknown conformations, objects in MIL are represented by groups of instances called \textit{bags} where each instance is described by its own feature vector, instead of represented by a single feature vector as in standard supervised learning.
Because only molecule-level drug binding suitability is known to human experts, MIL algorithms are only coarsely supervised at the bag-level, while fine-grained instance labels are unknown.

Utilizing the bag-level supervision, the prediction tasks of MIL are two-fold: predicting the bag labels, e.g., whether a molecule is suitable for making drugs or a human organ contains cancerous cells; 
and predicting the instance labels, e.g., which specific molecular conformations are suitable or which particular cells are cancerous.
Although theoretical results have demonstrated the feasibility of learning accurate instance concepts from bag labels \cite{Doran2016}, the empirical performances of weakly supervised instance label predictions are far behind their fully supervised counterparts. 

The inexactness in MIL supervision causes the main difficulty of instance label prediction \cite{Zhou2017}.
Under the standard multi-instance assumption \cite{Foulds2010}, the bag labels only inform the learner that instances from negative bags are negative, 
while the exact labels for instances in the positive bags are unknown.
Most previous instance-level MIL algorithms solve the above problem by disambiguating the bag labels, i.e., finding the most positive instance from positive bags \cite{Jan2000,Wang2018} or utilizing the attention mechanism to infer how the different instances contribute to the bag label \cite{Ilse2018,Shi2020,Rymarczyk2021}.

\begin{figure}[!t]
	\centering
	\includegraphics[width=\linewidth]{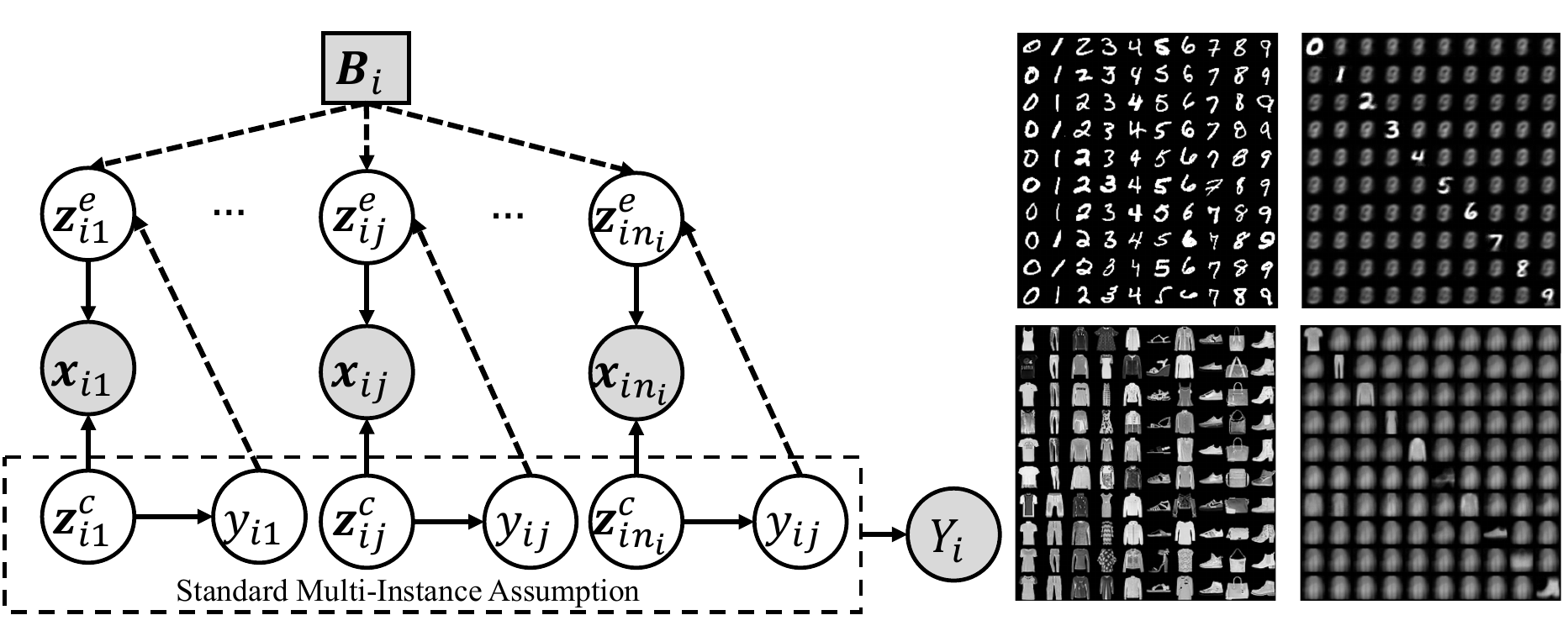}
	\caption{(Left) Our proposed graphical model. Shaded nodes denote observed variables and white nodes represent the latent ones. Dashed lines denote edges that may exist depending on the bags, while solid lines indicate relationships that are invariant with respect to the bags. 
	(Right) Multi-instance bags of MNIST and FashionMNIST images and their reconstructions from $\bm{z}^c$  (each class per row) inferred by CausalMIL.
		%		(These figures are best viewed when zoomed)
	}
	\label{model}
\end{figure}
%\begin{figure}[!t]
%	\centering
%	\begin{subfigure}{0.6\textwidth}
%	\centering
%	\includegraphics[width=\linewidth]{figures/figure.pdf}
%	\caption{}
%	\end{subfigure}
%	\begin{subfigure}{0.23\textwidth}
%		\centering
%		\includegraphics[height=1.2in]{figures/MNIST_original_grid.png}
%		\caption{}
%	\end{subfigure}
%	\begin{subfigure}{0.23\textwidth}
%		\centering
%		\includegraphics[height=1.2in]{figures/MNIST_recon_grid.png}
%		\caption{}
%	\end{subfigure}
%	\begin{subfigure}{0.23\textwidth}
%		\centering
%		\includegraphics[height=1.2in]{figures/Fashion_original_grid.png}
%		\caption{}	
%	\end{subfigure}
%	\begin{subfigure}{0.23\textwidth}
%		\centering
%		\includegraphics[height=1.2in]{figures/Fashion_reconstruct_grid.png}
%		\caption{}	
%	\end{subfigure}
%	\caption{(a) MNIST digits. (b) Reconstructions of each target digit per row by CausalMIL under bag-level supervision. (c) FashionMNIST items. (b) Reconstructions of each target fashion item per row. 
%%		(These figures are best viewed when zoomed)
%	}
%	\label{model}
%\end{figure}

%In this paper, we argue that multi-instance bags and their inexact bag-level supervision contain valuable information that can be utilized for learning invariant causal representations. 
% and their inexact bag-level supervision 
%\cite{Schott2018,Sun2021}

In this work, we propose to utilize the information hidden in the multi-instance bags for identifying the latent factors that generate the observed instances, and utilize the bag labels for encouraging disentanglement among the latent factors.
Specifically, we consider each observed instance as generated from instance-specific causal factors $\bm{z}^c_{ij}$ and bag-inherited non-causal factors $\bm{z}^e_{ij}$ as depicted in Figure \ref{model} (left). 
The instance-specific factors $\bm{z}^c_{ij}$ are responsible for capturing information that is causal to the instance labels, while the bag-inherited factors $\bm{z}^e_{ij}$ capture spurious information that may be inherited from the bags and correlates with the instance labels.
The bag labels are considered as the effects of their instance labels according to the standard multi-instance assumption.

By identifying and disentangling the latent factors, our model brings two advantages over existing MIL algorithms. 
Firstly, as the identified latent factors capture semantic meaningful causal representations are of much lower dimensions than the observed instances, their identification significantly reduces the difficulties of inferring instance labels from bag supervision.
%Firstly, as non-causal information of the labels such as styles and noises are captured in the bag-related factors, their identification and disentanglement significantly reduces the difficulty of instance label prediction.
%Since the disentanglement relies on the auxiliary information provided by the instances within the same bag, our approach mitigates the ambiguity of bag-level supervision.
Secondly, since only the instance-specific factors $\bm{z}^c_{ij}$ are causes to the instance labels $y_{ij}$, disentangling $\bm{z}^c_{ij}$ from $\bm{z}^e_{ij}$ remove the spurious correlations between $\bm{B}_i$ and $y_{ij}$. 
Therefore, identification and disentanglement of $\bm{z}^c_{ij}$ is not only beneficial for instance label predictions since $\bm{z}^e_{ij}$ will vary among bags, but is also useful for out-of-distribution (OOD) generalization since the factors that are the effects of instances labels are excluded from the predictive model \cite{Arjovsky2019,Ahuja2020,lu2022}.
It is worth noting that although $\bm{B}_i$ and $y_{ij}$ are unconditionally independent, they become conditionally dependent because previous algorithms that do not identify or disentangle the latent factors implicitly condition on either $\bm{x}_{ij}$ or $\bm{z}^e_{ij}$ (i.e., the collider or descendants of the collider in a v-structure \cite{Pearl2009}).

Our identifiability result builds upon the recent identifiable variational autoencoder (iVAE) \cite{Khemakhem2020}, but extends to a more flexible non-factorized prior distribution conditioning on the bag information instead of the factorized prior conditioning on a single instance assumed in the original iVAE. 
Furthermore, we propose to utilize a permutation invariant set transformation network for aggregating information from instances within the same bag to facilitate identifiability. 
Because the set transformation network universally approximates any set function \cite{Zaheer2017}, instances within the same bag are not only treated as sources of supervision inexactness but also effectively contribute to identifiability via conditioning the prior distribution.
We summarize our key contribution as follows:

%For example, in Figure \ref{illustration}, we show the original MNIST and FashionMNIST images along with their reconstructions from the latent factors learned by our proposed CausalMIL algorithm. In (b) and (d), each row contains reconstructions from the learned representations using one digit/item as the positive concept. 
%Firstly, we can see that the reconstructions capture the causal semantics of the concepts and are free of style variations. For example, the reconstruction of `9' is consisted of a circle at the top with a short line originating from the bottom, and the blurry line corresponds to the fact that its angle does not determine whether a digit is `9'.
%Secondly, we can see that only the one belonging to the positive concept is reconstructed to its semantic meaning, while the others are ignored.

\begin{itemize}
	\item We propose a novel CausalMIL algorithm that utilizes the bag information with permutation invariant set transformation networks to theoretical guarantee the identifiability of latent factors under non-factorized conditional exponential family priors;
	
	\item We propose a general graphical model covering many real-world MIL applications, and show that the causal and non-causal factors of the instance labels can be disentangled from the identified latents based on bag-level supervision.
%	\item We propose a targeted approach for disentangling the causal and non-causal latent factors of the instance labels using only bag-level supervision, and discuss how to utilized the disentangled factors for instance label prediction and out-of-distribution generalization.
	\item  Using a variety of datasets, we show that CausalMIL significantly outperforms existing MIL algorithms on instance label prediction tasks, and achieves better out-of-distribution generalization capabilities when compared to supervised invariant algorithms.
	
\end{itemize}

\section{Preliminaries}
\subsection{Notations}
Let $\mathcal{X} = \mathbb{R}^m$ denote the instance space and $\mathcal{Y} = \{0,1\}$ denote the label space. The training data contains samples organized in $n$ training bags $\mathcal{B} = \{\mathbf{B}_1, \cdots, \mathbf{B}_i,\cdots, \mathbf{B}_n \}$, where each bag is a set that contains different numbers of $n_i$ instances, i.e., $\mathbf{B}_i = \{\bm{x}_{i1}, \cdots, \bm{x}_{ij}, \cdots, \bm{x}_{in_i} \}$ with $\bm{x}_{ij} \in \mathcal{X}$. In the rest of this paper, we will also use $\mathbf{B}_i$ to denote the auxiliary information contained in the bag by abuse of notation.
The learning algorithm is provided with bags $\mathbf{B}_i$ and their associated bag labels $Y_i \in \mathcal{Y}$ during training. 
%For each instance $\bm{x}_{ij}$, there also exists an instance label $y_{ij} \in \mathcal{Y}$; however, the instance labels are \emph{unknown} to the learners.
Although there exists instance labels $y_{ij} \in \mathcal{Y}$ for each $\bm{x}_{ij}$, the instance labels are \emph{unknown} to the learners. 
We follow the standard multi-instance assumption \cite{Foulds2010} which assumes that positive bags contain at least one positive instance, and negative bags contain only negative instances.
It is worth noting that the assumption only requires bags to be i.i.d., the instances within the same bag can be dependent.
%In the rest of this paper, we will also refer to the positive instances as the \textit{target} instances and denote them as $\bm{x}_t$ as they are our target for learning the latent representations.

\subsection{VAE and Identifiability}
We briefly introduce Variational Autoencoder (VAE) and its identifiability results.
VAE can be considered as the combination of a generative latent variable model and an associated inference model, where both models are parameterized by neural networks \cite{Kingma2014}.
Specifically, VAE learns the joint distribution $p_{\bm{\theta}}(\bm{x},\bm{z}) = p_{\bm{\theta}}(\bm{x}\vert \bm{z} ) p_{\bm{\theta}}(\bm{z})$ where $p_{\bm{\theta}}(\bm{x}\vert \bm{z} )$ is the conditional distribution of observing $\bm{x}$ given $\bm{z}$, 
$\bm{\theta}$ is the set of generative parameters, 
and $p_{\bm{\theta}}(\bm{z}) = \Pi_{i=1}^d p_{\bm{\theta}}(z_i)$ is the factorized prior distribution of the latents.
By introducing an inference model $q_{\bm{\phi}}(\bm{z}\vert \bm{x})$,
the set of parameters $\bm{\phi}$ and $\bm{\theta}$ can be jointly optimized through maximizing the evidence lower bound (ELBO) on the marginal likelihood $p_{\bm{\theta}}(\bm{x} )$:
\begin{align}
		\mathcal{L} &= \mathbb{E}_{q_{\bm{\phi}}(\bm{z}\vert\bm{x})} [\log p_{\bm{\theta}}(\bm{x}\vert \bm{z})] - 
		D_{\text{KL}}(q_{\bm{\phi}} (\bm{z}\vert \bm{x}) \vert\vert p(\bm{z}) ) \nonumber\\
		&= \log p_{\bm{\theta}}(\bm{x}) - D_{\text{KL}} (q_{\bm{\phi}}(\bm{z}\vert\bm{x}) \vert\vert p_{\bm{\theta}}(\bm{z}\vert\bm{x}))
		\leq \log p_{\bm{\theta}}(\bm{x} ), 
%		\nonumber
	\end{align}
where $D_{\text{KL}}$ denotes the KL-divergence between the approximation and the true posterior, and $\mathcal{L}$ is a lower bound of the marginal likelihood $p_{\bm{\theta}}(\bm{x} )$ because of the non-negativity of the KL-divergence. 

Recently, it has been shown that VAEs with unconditional prior distributions $p_{\bm{\theta}} (\bm{z})$ are not identifiable \cite{Locatello2019}, but the latent factors $\bm{z}$ can be identified with a conditionally factorized prior distribution  $p_{\bm{\theta}}(\bm{x}|\bm{u})$ over the latent variables to break the symmetry, where $\bm{u}$ is an additionally observed variable \cite{Khemakhem2020}.
%Here we slightly abuse the notation by dropping the per data point summation in order to avoid cluttering. However, it is clear that the likelihood is calculated as the sum of ELBOs for each observed data sample \cite{Kingma2014}.

\newtheorem{assumption}{Assumption}
\section{Methods}
\subsection{CausalMIL}
We now formally describe our graphical model depicted in Figure \ref{model} (left). For any given bag $\bm{B}_i$, we model its observed instances $\bm{x}_{ij}$ as generated from unknown latent factors $\bm{z}_{ij} = (\bm{z}^c_{ij}, \bm{z}^e_{ij}) \in \mathbb{R}^d$ with $d \ll m$. 
We let $\bm{z}^c_{ij}$ correspond to the causes of the instance label $y_{ij}$, i.e., the line stroke that makes a digit represent the Arabic number `1' in Figure \ref{model} (right). In contrast $\bm{z}^e_{ij}$ captures other factors related to the observed $\bm{x}$ but are not causes of the instance label, i.e., the angle, thickness, and color of the line stroke that are not causal to `1'. 
The solid lines of the graphical model indicate that the causal relationships are invariant across all bags, and the dashed lines denote that their relationships may vary or even be absent across different bags.
%The unobserved instance label $y_{ij}$ is the effect of causal factors $\bm{z}^c_{ij}$, and also is the cause to non-causal factors $\bm{z}^e_{ij}$. 
The bag label $Y_i$ is determined by the labels of its instances $y_{ij}$ according to the standard multi-instance assumption.
To sum up, we posit that the graphical structure in Figure \ref{model} should satisfy the following assumptions:

\begin{assumption}
(a) The bag labels are determined by its instances according to the standard multi-instance assumption;
(b) the causal graph in Figure \ref{model} is a Directed Acyclic Graph (DAG); 
(c) given the latent factors $\bm{z}_{ij}$, the observed instances are independent of the labels and the bags: $\bm{x}_{ij} \indep y_{ij},Y_i,\bm{B}_i | \bm{z}_{ij}$, i.e., the generation mechanism $p(\bm{x}_{ij} | \bm{z}_{ij})$ is invariant across bags; 
(d) $y_{ij} \indep \bm{B}_i \vert \bm{z}^c_{ij}$, i.e., the mechanism between the causal factors $\bm{z}^c_{ij}$ and $y_{ij}$ is invariant across bags.

\label{assumption}
\end{assumption}
%This assumption is more general than the factorized conditional distribution assumed by iVAE because Figure \ref{model} allows for arbitrary connections among the components within $\bm{z}^c_{ij}$  $\bm{z}^e_{ij}$ as long as the model remains a valid DAG, i.e., there exists no directed cycles.
%Instead, we only require that there should be no connections among $\bm{z}^c_{ij}$ and $\bm{z}^e_{ij}$ given $y_{ij}$, which makes our model in Figure \ref{model} more practical for real-world applications.

We now discuss the practicality of Assumption \ref{assumption}. 
Firstly, the standard multi-instance assumption commonly used in a wide range of MIL applications such as medical image analysis \cite{Li2021}, fine-grained sentiment analysis \cite{Angelidis2018}, and sound event detection \cite{Wang2019}. 
Secondly, modeling the causal structures as DAGs are common in causal discovery \cite{Pearl2009,Peters2016}. To the best of our knowledge, most of the causality-based machine learning literature assumes the causal structure to be acyclic.
Thirdly, it also makes sense that generating mechanism from latent factors to observed instances $p(\bm{x}_{ij} | \bm{z}_{ij})$ is invariant since causal mechanisms are considered to be stable across environments \cite{Scholkopf2021}. Otherwise, it would be impossible to infer $\bm{z}$ from $\bm{x}$ for any test instances.
Lastly, the fourth assumption is not only commonly adopted in invariant and causal representation learning literature \cite{Arjovsky2019,Ahuja2020,lu2022}, but also suitable for MIL applications. For example, when diagnosing whether a cell from a tissue is cancerous, the causal mechanism of diagnosing a cell should be invariant across bags, while the non-causal factors caused by the patient or equipment differences may often change \cite{Zhang2021}.
%In fact, the causal structure and assumptions in Figure \ref{model} and Assumption \ref{assumption} covers most scenarios of MIL applications. For example, in weakly supervised medical image classification \cite{Skrede2020,Lu2021} the task is to identify cancerous cells (instance) from an organ (bag) with only organ-level labels

From now on we will drop the subscript to avoid cluttering the notations. In the following discussion we will show that when the underlying data generating mechanism satisfies Assumption \ref{assumption}, the latent variables $\bm{z}$ can be identified up to permutation and affine transformations if the conditional prior distribution $p(\bm{z} \vert \bm{B}, y)$ belongs to a general exponential family distribution:
\begin{assumption}
The prior distribution of the latent factors $p(\bm{z} \vert \bm{B}, y)$ follows a general exponential family with its parameter specified by an arbitrary function $\bm{\lambda} (\bm{B}, y)$ and sufficient statistics $\bm{T} (\bm{z}) = [\bm{T}_f(\bm{z})^T, \bm{T}_{NN}(\bm{z})^T ]^T$. Here the sufficient statistics $\bm{T} (\bm{z})$ is defined by the concatenation of $\bm{T}_f(\bm{z})=[\bm{T}_1(z_1)^T,\cdots,\bm{T}_d(z_d)^T]^T$ from a factorized exponential family and the outputs of a neural networks $\bm{T}_{NN}(\bm{z})$ with universal approximation power.
Then, the probability density can be written as:
\begin{align}
	p_{\mathbf{T},\bm{\lambda}} (\bm{z}\vert \bm{B}, y ) = 	\frac{ \mathcal{Q}(\bm{z})}{ \mathcal{C}(\bm{B}, y)} \exp [\bm{T} (\bm{z})^T \bm{\lambda} ( \bm{B} , y)],
\label{nf-exponential}
\end{align}
where $\mathcal{Q} (\bm{z})$ is the base measure and $\mathcal{C}(\bm{B}, y)$ is the normalizing constant.
\label{exponential}
\end{assumption}

The above assumption is inspired by the recent advancement of iVAE \cite{Khemakhem2020}. However, there are two differences. 
%between the above non-factorized conditional prior $p(\bm{z}\vert\bm{B}, y)$ and the factorized exponential family distribution assumed by iVAE \cite{Khemakhem2020}. 
Firstly, because Figure \ref{model} allows for connections among the components $\bm{z}_{ij}$ within $\bm{z}$ as long as the generative model remains a valid DAG, our non-factorized conditional prior assumption is more general than the factorized distribution assumed by iVAE.
In Equation \ref{nf-exponential}, we utilize a neural networks $\bm{T}_{NN}(\bm{z}^c)$ for capturing the interactions among components of $\bm{z}$. 
%However, it is worth noting that $\bm{z}^c$ and $\bm{z}^e$ are independent since $\bm{T}_{NN_1}$ and $\bm{T}_{NN_2}$ are individually parameterized.
Secondly, our additionally observed auxiliary information is provided by multi-instance bag $\bm{B}$ which may consist arbitrary number of instances instead of a single scalar or vector-valued instance $\bm{u}$ as used by iVAE. 
Therefore, it is natural to ask (1) if identifiability still holds for non-factorized priors, and (2) how to aggregate information from bags of varying sizes for breaking the symmetry of the prior distribution?

To answer the above questions, we first describe our generative model and discuss its identifiability. Formally, let us consider the following generative models under Assumption \ref{nf-exponential}:
\begin{align}
	p_\mathbf{\bm{\theta}} (\bm{x},\bm{z}|\bm{B}, y) &= p_{\bm{f}} (\bm{x}\vert \bm{z}) p_{\mathbf{T},\bm{\lambda}} (\bm{z}\vert \bm{B}, y),
	\label{eq_generative}\\
	p_{\bm{f}} (\bm{x}\vert \bm{z}) &= p_{\bm{\varepsilon}}(\bm{x} - f(\bm{z})) \label{eq_additive_noise},
\end{align}
where Equation \ref{eq_generative} describes the generative process of $\bm{x}$ given the underlying latent factors $\bm{z}$, along with the bag context $\bm{B}$ and the instance label $y$.
Equation \ref{eq_additive_noise} implies that the observed representation $\bm{x}$ is an additive noise function, i.e., $\bm{x}=f(\bm{z}) + \bm{\varepsilon}$ where $\bm{\varepsilon}$ is independent of $\bm{f}$ or $\bm{z}$. 

Following standard VAE and iVAE derivation, the evidence lower bound for each instance $\bm{x}$ in bag $\bm{B}$ of the above generative model can be written as:
\begin{align}
	%	\mathcal{L}_{\text{ELBO}} =   \mathbb{E}_{q_{\bm{\phi}}(\bm{z}_t \vert \bm{x}_t)}  [\log p_{\bm{\theta}} (\bm{x}_t\vert \bm{z}_t)] - D_{KL} (q_\phi (\bm{z}_t\vert \bm{x}_t) \vert\vert p(\bm{z}_t \vert \bm{B}) )
	\mathcal{L}_{\text{ELBO}} &=   \mathbb{E}_{q_{\bm{\phi}}(\bm{z} \vert \bm{x},\bm{B},y)}  [\log p_{\bm{f}} (\bm{x}\vert \bm{z}) + \log p_{\bm{T,\lambda}}(\bm{z} \vert \bm{B}, y) \\
	& - \log q_{\bm\phi} (\bm{z} \vert \bm{x},\bm{B},y)] 
	+ \log p(\bm{B}).
	\label{ELBO-MIL}
\end{align}
Note that the ELBO in Equation \ref{ELBO-MIL} contains an additional term of the bag prior $\log p(\bm{B})$ that do not affect identifiability but improves the estimation for the conditional prior \cite{Mita2021}.
Our main identifiability results can now be described as:
\begin{theorem}
\label{thm-identifiability}
Assume we observe instances from multi-instance bags sampled according to Equation \ref{nf-exponential}-\ref{eq_additive_noise}.
% where $p_{\bm{f}}(\bm{x} \vert \bm{z})$ has additive noise as described in Equation \ref{eq_additive_noise}, and $p_{\bm{T},\bm{\eta}}(\bm{z}\vert \bm{B}, y)$ is an exponential family distribution with parameters $\bm{\theta}=(\bm{f},\bm{T}, \bm{\lambda})$ as specified in Equation \ref{nf-exponential}, 
Furthermore, assume that the following holds:
\begin{enumerate}[i.]
	\item The set $\{\bm{x} \in \mathcal{X}: \varphi_{\bm{\varepsilon}}(\bm{x}=0\}$ has measure zero, where $\varphi_{\bm{\varepsilon}}$ is the characteristic function of the density $p_{\bm{\varepsilon}}$ defined in Equation \ref{eq_additive_noise}.
	\item The function $\bm{f}$ is injective and all of its second-order cross partial derivatives exist.
%	\item The function $\bm{f}$ is injective.
	\item The sufficient statistics $T_{\bm{f}}$ are twice differentiable.
	\item There exist $k+1$ distinct bags $(\bm{B}^0,y^0), \dots, (\bm{B}^{d}, y^d)$ such that the $k\times k$ matrix $E$ is invertible, where $k$ is the dimension of $\bm{T}$ and $E$ is defined as:
	\begin{equation}
		E= (\bm{\lambda}(\bm{B}^1, y^1) - \bm{\lambda}(\bm{B}^0,y^0), \dots, \bm{\lambda}(\bm{B}^{k}, y^k) - \bm{\lambda}(\bm{B}^0, y^0) )
	\end{equation}
\end{enumerate}
Then, the parameters $\bm{\theta}=(\bm{f},\bm{T}, \bm{\lambda})$ are identifiable up to an equivalence class induced by permutation and componentwise transformations. 
%Specifically, if $p_{\bm{\theta}} (\bm{x},\bm{z}|\bm{B}, y) =  p_{\bm{\theta}^\prime} (\bm{x},\bm{z}|\bm{B}, y)$ where $\bm{\theta}^\prime$ is another set of parameters in the class, we have the following equivalence between $\bm{\theta}$ and $\bm{\theta}^\prime$: for any $\bm{x}_t \in \mathcal{X}$, $\exists \bm{A},\bm{c}$ such that
%\begin{equation}
%	T(\bm{f}^{-1} (\bm{x}_t)) = \bm{A} \bm{T}^{\prime} (\bm{f}^{\prime-1} (\bm{x}_t)) + \bm{c},
%\end{equation}
%where $\bm{A}$ is an invertible matrix of $kd \times kd$ dimension and $\bm{c}$ is a vector of dimension $kd$.
\label{identifiability}
\end{theorem}

%Theorem \ref{identifiability} extends upon the main result of iVAE in the sense that our results build the identifiability upon a non-factorized prior distribution, which is more flexible and general than the conditionally factorized prior used in iVAE. 

It is interesting to observe that condition (iv) of Theorem \ref{identifiability} is easier to satisfy in MIL than standard supervised learning. 
Intuitively, the condition requires that the auxiliary information should ``break the symmetry'' in the representation space the model could learn. 
An intuitive analogy is inferring an object's shape from its shadow: 
if we only observe one shadow of the object, it's difficult to know its shape; 
however, if we observe multiple objects under similar lighting conditions (from a bag of instances), we may identify the lighting.
If we observe an object under different lighting conditions (from many bags of instances), we may identify the underlying shape. 
Another analogy is to consider the bags as H\&E stained histopathology images and the instances as image patches: in one bag, we observe cells under the same staining process, which provides information regarding the staining procedure; 
in different bags, we observe cancerous cells under different staining procedures, which makes it possible to infer the causal representations of cancerous cells. 
In standard supervised learning, instances are independent with no auxiliary information given. 
In MIL, instances within bags are naturally dependent and organized in bags, which may then provide the necessary auxiliary information for latent identifiability. 
We provide the proof for Theorem \ref{identifiability} in the Appendices.
%Such information is available in MIL but not in a standard learning setting. 

Technically, condition (iv) can be restated as requiring the vectors $(\bm{\lambda}(\bm{B}^1, y^1) - \bm{\lambda}(\bm{B}^0,y^0), \dots, \bm{\lambda}(\bm{B}^{k}, y^k) - \bm{\lambda}(\bm{B}^0, y^0) )$ to be independent. Therefore, the instance label $y$ becomes unnecessary if there exist $k+1$ distinct bags such that $E=(\bm{\lambda}(\bm{B}^1) - \bm{\lambda}(\bm{B}^0), \dots, \bm{\lambda}(\bm{B}^{k}) - \bm{\lambda}(\bm{B}^0) )$ of size $k\times k$ is invertible,
which is especially attractive in MIL where the instance labels are not available but the bags are abundant. 
%In other words, (iv) is usually satisfied when the bag auxiliary information is diverse.

%In practice we choose $p(\bm{z} \vert \bm{B}, y)$ to be a non-factorized Gaussian distributions whose location and scale is conditioned on $\bm{B}$, which is a special case of the exponential family distributions with two dimensional sufficient statistics. 
%The non-factorized prior extends the results of the original iVAE where the prior is restricted to factorized distributions.

Two practical issues need to be addressed before applying the identifiability theorems to solve MIL problems.
The first hurdle lies in how to characterize the auxiliary information for conditioning the prior distribution of $\bm{z}$.
This is different from standard iVAE where the auxiliary information is readily provided in a single-instance $\bm{u}$ since the auxiliary information in MIL are provided in the form of instances sampled from bags.
To make things worse, the auxiliary information also cannot be derived from the bag labels or the multi-instance assumption. 
This is because they only tell us the bag labels and the relationship between instances labels and bag labels, instead of the generative process of the bags and their instances.

To solve the above problem, we propose to utilize a trainable permutation invariant function parameterized by deep neural networks. Formally, we define:
\begin{align}
	\bm{B}_i = net(\{\bm{x}_{i1}, \cdots, \bm{x}_{in_i} \}) = \rho[pool(\{ \phi(\bm{x}_{i1}), \cdots, \phi(\bm{x}_{in_i}) \})],
	\label{set_pooling}
\end{align}
where $\rho$ and $\phi$ are arbitrary continuous functions implemented through neural networks, $pool$ is the sum operator over the instances $\bm{x}$ transformed by the neural network $\phi$, and finally $\rho$ is another neural network that operates on the pooled transformations. 
An important property of the transformation defined in Equation \ref{set_pooling} is that it is capable to express any permutation invariant function that operates on set inputs \cite{Zaheer2017}. 
Since a multi-instance bags are both permutation invariant and contain instances of varying sizes \cite{Ilse2018}, it is adequate to model its bag information as a set function of the instances.
%Therefore, Equation \ref{set_pooling} is suitable candidate for inferring the bag-level factor from its instances. 
%To the best of our knowledge, this is the first application of deep set learning in MIL.

The second missing piece of the puzzle is how to separate $\bm{z}^c$ from $\bm{z}^e$ in the identified latent factors $\bm{z}$, since we would only require $\bm{z}^c$ for the downstream tasks. 
To solve this problem, we take a different approach from a typical VAE-based disentanglement algorithm where their goal is to ensure the components of $\bm{z}$, i.e., $z_p$ for $p=1,\cdots,d$, are independent and correspond to meaningful semantics \cite{Higgins2017}. 
Although component-wise disentanglement is desirable for causal representation learning in general, due to the fact that multi-instance bags are only weakly supervised at the bag level, we argue that a more pragmatic approach is to utilize the bag labels for separating the causal factors $\bm{z}^c$ of the instance labels from the non-causal ones $\bm{z}^e$ without requiring additional information.

We now revisit the graphical model in Figure \ref{model} to discuss the benefit of not pursuing component-wise disentanglement and using only $\bm{z}^c$ for instance classification. 
The instance labels $y$ are correlated with both $\bm{z}^c$ and $\bm{z}^e$, and thus the entirety of $\bm{z}$ is predictive of $y$.
Therefore, for supervised learning tasks that assume i.i.d. training and test data, using $\bm{z}$ is preferable for prediction.
However, in MIL our goal is to predict the instance labels from unseen test bags. Since the bag-inherited factor $\bm{z}^e$ may be different among training and test bags, using $\bm{z}^e$ would negatively affect prediction.
Furthermore, since instance classifiers in MIL must infer instance-level labels from bag-level weak supervision, excluding spurious correlations and including only the low dimensional causal representation also reduces the difficulty in learning the instance classifier.
Another important benefit of separating $\bm{z}^c$ from $\bm{z}^e$ is that it also promotes OOD generalization. By considering bags as environments and the non-causal factors $\bm{z}^e$ as the effects of environment-induced biases, using $\bm{z}^c$ and excluding $\bm{z}^e$ improves prediction robustness since $p(y\vert \bm{z}^c)$ is invariant across bags as specified by Assumption \ref{assumption}(d), which also coincides with the goal of invariant causal prediction algorithms such as \cite{Arjovsky2019,Ahuja2020,lu2022}.

If the instance labels are known, then our goal of disentanglement can be straightforwardly achieved by utilizing the PC algorithm \cite{Spirtes2000} to learn a Markov equivalent class of DAGs and find the direct causes as done in \cite{lu2022}.
Unfortunately, this is not the case for MIL where only bag labels are known.
To solve this problem, we propose to utilize the bag label $Y$ with the ELBO in Equation \ref{ELBO-MIL} using a bag-wise maximum operator. Specifically, we optimize for the following target
%\begin{align}
%	\mathcal{L}_{\text{CausalMIL}} & =  \max\limits_{\bm{z} \in \bm{B}}
%%	 \mathbb{E}_{q_{\bm{\phi}}(\bm{z} \vert \bm{x},\bm{B},Y)} 
%	 \{\log p_{\bm{f}} (\bm{x}\vert \bm{z}) 
%	+ \alpha\log p_{\bm{\omega}}(Y \vert \bm{z})    
%	- \text{KL}  [ q_{\bm{\phi}} (\bm{z} \vert \bm{x},\bm{B}) || p_{\bm{T,\lambda}}(\bm{z} \vert \bm{B},Y) ]\} \nonumber\\
%	&
%	+ \log p_{\bm{\vartheta}} (\bm{B}\vert \bm{z}) 
%	- \text{KL} [q_{\bm{\psi}}(\bm{z} \vert \bm{B}) \vert\vert p(\bm{B}) ] 
%%	+ \log p_{\bm{T,\lambda}}(\bm{z}^c, \bm{z}^t \vert \bm{B},Y) \\
%%	&- \log q_{\bm\phi} (\bm{z}^c \vert \bm{x},\bm{B},Y)],
%\label{max_elbo}
%\end{align}
\begin{align}
	\centering
	\mathcal{L}_{\text{CausalMIL}} & =  
	\log p_{\bm{f}} (\bm{x}^{\ast}\vert \bm{z}^{\ast})  
	+ \alpha\log p_{\bm{\omega}}(Y \vert \bm{z}^{\ast}) 
	- \text{KL} [ q_{\bm{\phi}} (\bm{z}^{\ast} \vert \bm{x}^{\ast},\bm{B}) || p_{\bm{T,\lambda}}(\bm{z}^{\ast} \vert \bm{B}) ] \nonumber\\
	&
	+ \log p_{\bm{\vartheta}} (\bm{B}\vert \bm{z})
	- \text{KL} [q_{\bm{\psi}}(\bm{z} \vert \bm{B}) \vert\vert p(\bm{B}) ],
	\label{max_elbo}
\end{align}
where $\bm{z}^{\ast} = \arg\max_{\bm{B}} p_{\bm{\omega}} ( Y \vert \bm{z}) $, and $p_{\bm{\omega}} ( Y \vert \bm{z})$ is a \textit{linear} classifier with $\alpha$ as its weighting hyper-parameter. It is also worth noting that the above ELBO is written with respect to a multi-instance bag instead of the instances. Furthermore, the last two terms can be obtained by deriving an ELBO for $\log p_{\bm{\vartheta}}(\bm{B})$ in Equation \ref{ELBO-MIL} \cite{Mita2021}.

In other words, the reconstruction, classification, and conditional KL terms in the ELBO are optimized for \emph{only one instance per bag}, and the instance is chosen using a maximum operator to satisfy the standard multi-instance assumption.
%however, all instances are utilized using Equation \ref{set_pooling} to construct the conditional bag prior and to ensure identifiability. 
The formulation of Equation \ref{max_elbo} is effective for learning $\bm{z}^c$ while ignoring $\bm{z}^e$ because of the following two reasons.
Firstly, as the ELBO is optimized per mini-batches consisted of many multi-instance bags using the reparameterization trick \cite{Kingma2014}, the optimization process will force the linear classifier $p_{\bm{\omega}}  (y\vert \bm{z})$ to predict using the $\bm{z}^c$ and ignore $\bm{z}^e$; otherwise, the linearity of $p_{\bm{\omega}} (y\vert \bm{z})$ will not be able to predict the instance labels since $\bm{z}^e$ are different for different bags in the mini-batch. 
Moreover, because the reconstruction probabilities are evaluated at only one latent factor for each bag, the encoder $q_{\bm{\phi}} (\bm{z}\vert \bm{x}, \bm{B})$ is encouraged to encode only $\bm{z}^c$ since it carries the content causal information for reconstructing the positive $\bm{x}$ and remains invariant across different bags in the mini-batch. 
To see this, consider the opposite scenario where the reconstruction losses are evaluated for every instance in the bag, the encoder would also encode $\bm{z}^e$ since it remains the same within that bag.
Therefore, CausalMIL is targeted to learn $\bm{z}^c$ that are causal to the positive instances of a bag, and ignores both the non-causal factors $\bm{z}^e$ and the causal factors for negative instances, as illustrated by the reconstructions in Figure \ref{model} (right) and Figure \ref{Fashion-details}.
%The essence of this idea is inspired by Targeted Maximum Likelihood Estimation for counterfactual causal inference \cite{Laan2006} where their estimation is targeted to only one particular parameter of interest and consider the remaining parameters as nuisances.
% which also serves as the namesake of CausalMIL.

%The maximum operator of Equation \ref{max_elbo} is fundamentally different from the pooling operators used in previous neural network-based MIL algorithms \cite{Jan2000,Wang2018} where they place max-pooling either at the last layer or before the fully connected layer of a feed-forward network; 
%therefore, the main purpose of their max-pooling operation is to simply \textit{select the most positive instance}.
%As a result, previous algorithms often exhibit high precision but low recall when predicting instance labels. However, the maximum operator in Equation \ref{max_elbo} is used to simultaneously disentangle $\bm{z}^c$ from $\bm{z}^e$ and to incorporate the standard multi-instance assumption, which makes CausalMIL effective not only at instance label prediction tasks but also at out-of-distribution generalizations.

%The ELBO of CausalMIL can be optimized using stochastic gradient descent with the re-parameterization trick \cite{Kingma2014}.
%After training, instance labels can be obtained by first infer the latent representation using the encoder $q_{\bm{\phi}}(\mathbf{z}|\pmb{x})$ and then perform prediction with the linear classifier $f_{cls}(\bm{z})$ using $\bm{z}$.
It is worth noting that there exist two types of approximate inference in CausalMIL. The first one, common to almost all VAE-based algorithms, is the amortization gaps between the ELBOs and the marginal log-likelihoods caused by amortizing the variational parameters over the entire dataset, instead of optimizing for each training sample individually. 
The second one, specific to CausalMIL, is the approximation gap of only inferring the log-likelihood of one instances per each bag. 
However, as corroborated by our experiment results, these approximation behaviors does not prevent the success of CausalMIL in instance label prediction and OOD generalization.

%Moreover, previous results have shown that by incorporating more expressive approximate posteriors instead of factorized conditional Gaussians, the model learns a better distribution over the data. We leave the endeavors towards these directions under the setting of multi-instance learning for future works.

\section{Experiments}
In this section, we first evaluate the instance label prediction performances of CausalMIL against MIL baselines including mi-Net \cite{Wang2018}, Attention-based MIL (Attn-MIL) \cite{Ilse2018}, Kernel Self-Attention-based MIL (KSA-MIL) \cite{Rymarczyk2021}, Multi-Instance Variational Autoencoder (MIVAE) \cite{Zhang2021}. 
Then, we evaluate the out-of-distribution generalization ability of CausalMIL by comparing with supervised learning algorithms including ERM and its variants, Invariant Risk Minimization (IRM) \cite{Arjovsky2019}, IRM with Game Theory (IRM GAME) \cite{Ahuja2020}, Invariant Causal Representation Learning (iCaRL) \cite{lu2022}.

%The first category of baselines includes traditional multi-instance learning algorithms that can predict instance labels and deal with pre-computed feature representations, including mi-SVM \cite{Andrews2002} and Variational Gaussian Process MIL (VGPMIL) \cite{Haussmann2017}.

We implemented CausalMIL using PyTorch and conducted most of the experiments with a single NVIDIA RTX3090 GPU.
Detailed settings, parameters, and data for reproducing the results are provided in the Appendices. The implementation code is publicly available at \url{https://github.com/WeijiaZhang24/CausalMIL}.
%Training for CausalMIL are conducted for 200 epochs and parameters are selected according to the validation ELBO evaluated using 10\% of the training bags. 

\subsection{Instance Label Prediction}
%\subsection{Qualitative Evaluation}
%For inspecting the latent representation inferred by CausalMIL, we utilize two multi-instance datasets based on MNIST \cite{Lecun1998} and FashionMNIST \cite{Xiao2017} constructed similar to the setting of the multi-instance 20 Newsgroup datasets \cite{Zhou2009}.

\subsubsection{Datasets}
Our quantitative evaluation first utilizes 30 multi-instance classification datasets generated from the MNIST \cite{Lecun1998}, FashionMNIST \cite{Xiao2017}, and KuzushijiMNIST \cite{Clanuwat2018} datasets. 
%MNIST contains 60,000 training images of handwritten digits and 10,000 of test digits where each digit is represented by 28$\times$28 grayscale pixels. 
%In order to investigate the setting of predicting instance labels from bag level supervision,
The bag construction procedure is similar to the multi-instance bags generation from the 20 Newsgroup corpus \cite{Zhou2009}. 
Specifically, we create the multi-instance MNIST-bags such that each bag contains a number of images where its bag size is drawn from a Gaussian distribution with fixed mean and variance.
The bag is positive if it contains a target digit, e.g., `4', and negative if otherwise. Using 10 different digits as targets, we obtain 10 multi-instance MNIST-bag datasets. Similarly, we construct 10 FashionMNIST-bags and 10 KuzushijiMNIST-bags. 
%The mean bag size to $50$ with a standard deviation of $10$ and each positive bag contains about 10\% of positive instances.

We also evaluate CausalMIL on a hematoxylin and eosin (H\&E) stained Colon Cancer histopathology task \cite{Sirinukunwattana2016}.
Histopathology from whole-slide images is an important application of MIL since supervised predictions require pathologists to provide pixel-level annotations which is extremely expensive and time-consuming.
The dataset contains 100 images obtained from tissues of either normal or malignant patient tissues.
For each image bag, the instances are generated as patches $27\times 27$ pixels using markings of major nuclei for each cell. 
A total amount of 22,444 instances (\textasciitilde 220 instances per bag) are provided with ground truth instance labels, i.e. whether the cell is epithelial.
A bag is labeled positive if it contains at least one epithelial patch and negative if otherwise.

\begin{table*}[!t]
	\small
	\centering
	\caption{Results of instance label prediction performances on MNIST, FashionMNIST, and KuzushijiMNIST bags.
		%	Experiments for the compared methods are run 5 times for each digit/fashion object. 
		The reported means $\pm$ std are the macro averaged over 10 one-vs-rest datasets, respectively. }
	%	\resizebox{1\linewidth}{!}{
		\begin{tabular}{l | c c| c c  |c c}
			\hline
			&	\multicolumn{2}{c|}{MNIST Bags} & \multicolumn{2}{c}{FashionMNIST Bags} & \multicolumn{2}{c}{KuzushijiMNIST Bags} \\
			\hline
			& F-score & AUC-PR & F-score &  AUC-PR &  F-score &  AUC-PR \\
			\hline
			%			mi-SVM \\
			%			VGPMIL & All TBD \\
			%			\hline
			mi-NET & 0.595$\pm$.078 & 0.702$\pm$.165 & 0.251$\pm$.290& 0.329$\pm$.450  & 0.502$\pm$.106 & 0.616$\pm$.164 \\
			Attn-MIL & 0.712$\pm$.147 & 0.776$\pm$.222 & 0.398$\pm$.243 & 0.534$\pm$.144 & 0.638$\pm$.145 & 0.654$\pm$.209 \\ 
			KSA-MIL	& 0.775$\pm$.092  & 0.845$\pm$.122 &  0.545$\pm$.288 &  0.617$\pm$.360 & 0.707$\pm$.084 & 0.782$\pm$.116 \\
			MIVAE & 0.901$\pm$.035 & 0.921$\pm$.056 &  0.701$\pm$.257 & 0.733$\pm$.271 & 0.779$\pm$.147 & 0.838$\pm$.102\\
			\hline
			CausalMIL& \textbf{0.966$\pm$.018} & \textbf{0.981$\pm$.018} & \textbf{0.748$\pm$.206} & \textbf{0.822$\pm$.168} & \textbf{0.833$\pm$0.109} & \textbf{0.934$\pm$.075} \\
			\hline
%			& & 
		\end{tabular}
		\label{Quantitative-MNIST}
	\end{table*}

\subsubsection{Quantitative Results}
%We emphasize that all of the reported quantitative results are for instance label predictions from bag level supervision.
%In other words, all algorithms are trained using only bag labels, and the ground truth instance labels are used only for calculating the instance label prediction performance metrics on the test set.

For quantitative evaluation, we use  the area under the Precision-Recall curve (AUC-PR), instead of area under the ROC curve due to class imbalance at the instance level. For example, although the original MNIST datasets are balanced, only \textasciitilde5\% of the instances are positive in the MNIST-bags. 
We also report the f-scores of the compared methods. 
For the methods that directly output instance prediction such as mi-NET, MIVAE, and CausalMIL, the results are obtained from thresholding their predicted scores.
For methods based on the attention mechanism, such as Attn-MIL and KSA-MIL, the results are obtained by first normalizing the instance-to-bag attention weights to $[0,1]$ and applying the threshold.
As there is usually no instance label available in real application scenarios, we use $0.5$ uniformly as the classification threshold.

Table \ref{Quantitative-MNIST} shows the quantitative results on MNIST, FashionMNIST, and KuzushijiMNIST bags. 
CausalMIL performs significantly better than the compared methods on both metrics. 
%KSA-MIL achieves higher recall on FashionMNIST-bags; however, its precision is significantly lower. 
Note that the large standard deviations for FashionMNIST-bags are because some fashion objects are much more difficult to classify (e.g., distinguishing shirts from pullovers and coats). Furthermore, we can see that attention-based MIL algorithms exhibit significant performance variations even for MNIST-bags. A possible reason is that when the data satisfy the standard multi-instance assumption, learning a weighted average of the individual instance contributions is not efficient.
Among the compared algorithms, CausalMIL achieves the best performances with the lowest standard deviations.
%These results show that the representations learned by CausalMIL are semantically meaningful and useful for instance label prediction.

Table \ref{histopathology} shows the instance prediction performances of the compared algorithms on the Colon Cancer dataset. 
The result trends are similar to previous ones.
For discriminative deep learning- based algorithms, attention-pooling (Attn-MIL) performs better than max-pooling (mi-Net), while incorporating the self-attention mechanism (KSA-MIL) further improves the performances. 
A recent VAE-based generative MIL algorithm (MIVAE) performs better than the discriminative algorithms but still significantly worse than CausalMIL because it does not learn meaningful representations and does not provide identifiability guarantee. 
%(For comparisons of the learned latents between CausalMIL and MIVAE, please refer to the ablation results in the Appendix).

\begin{table}[!t]
	\centering
	\caption{Instance label prediction performances of the Colon Cancer dataset. Experiments are repeated for 5 times and the average metric$\pm$standard deviation of 5-fold cross validations are reported.}
	\begin{tabular}{l| c H H H c}
		\hline
		Method & F-score  & Precision & Recall & F-Score & AUC-PR \\
		\hline
		mi-Net & 0.392$\pm$.017 & 0.866$\pm$0.017 & 0.816$\pm$0.031 & 0.813$\pm$0.023 & 0.491$\pm$.028\\
		Attn-MIL & 0.466$\pm$.037 & 0.944$\pm$0.016& 0.851$\pm$0.035 & 0.893$\pm$0.022 & 0.536$\pm$.014\\
		KSA-MIL & 0.510$\pm$.029 & 0.944$\pm$0.016& 0.851$\pm$0.035 & 0.893$\pm$0.022 & 0.578$\pm$.025\\
		MIVAE&  0.675$\pm$.033 & & & &0.747$\pm$.032\\
		\hline
		\textbf{CausalMIL} & \textbf{0.833$\pm$.024} & & & & \textbf{0.878$\pm$.046} \\
		\hline
	\end{tabular}
	\label{histopathology}
\end{table}

\begin{figure*}[!t]
	\centering
	\begin{subfigure}{0.16\textwidth}
		\includegraphics[height=0.9in]{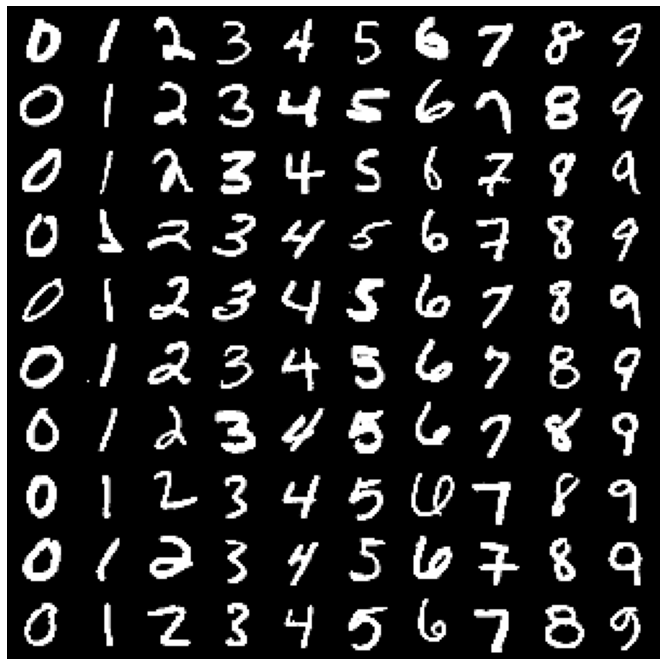}
		%		\caption{}
	\end{subfigure}
	\begin{subfigure}{0.16\textwidth}
		\includegraphics[height=0.9in]{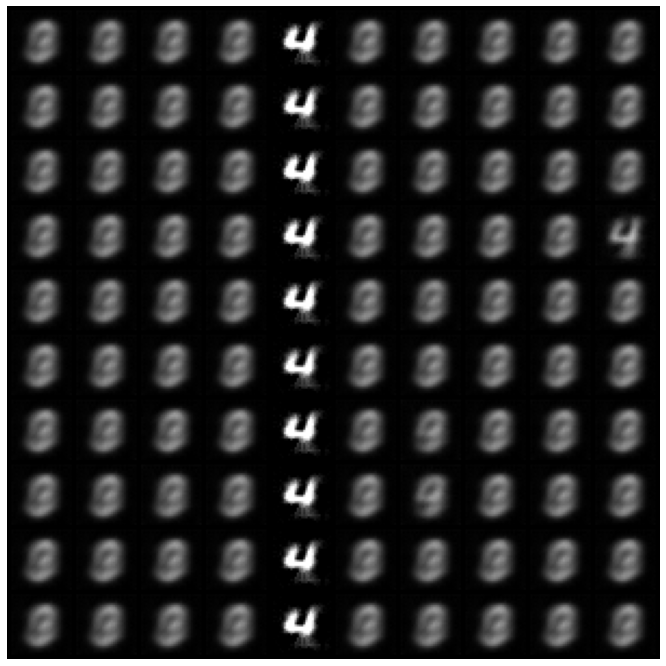}
		%		\caption{}
	\end{subfigure}
	\begin{subfigure}{0.16\textwidth}
		\includegraphics[height=0.9in]{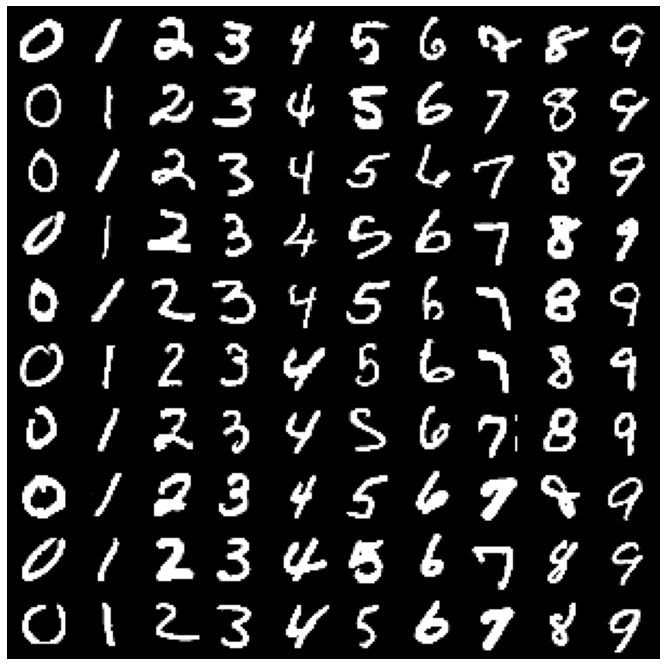}
		%		\caption{}
	\end{subfigure}
	\begin{subfigure}{0.16\textwidth}
		\includegraphics[height=0.9in]{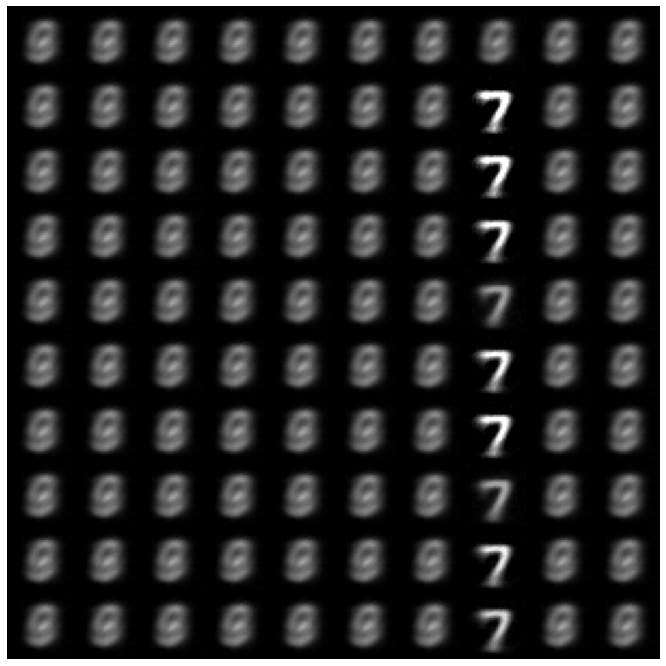}
		%		\caption{}
	\end{subfigure}
	\begin{subfigure}{0.16\textwidth}
		\includegraphics[height=0.9in]{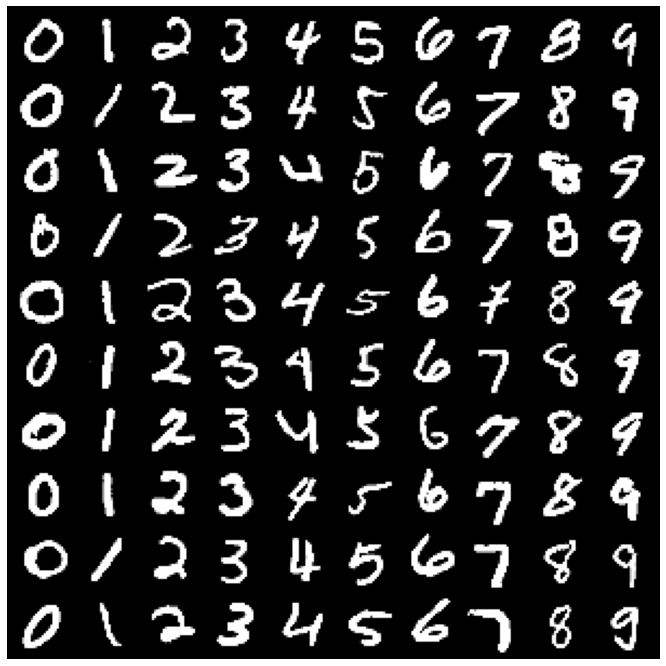}
		%		\caption{}
	\end{subfigure}
	\begin{subfigure}{0.16\textwidth}
		\includegraphics[height=0.9in]{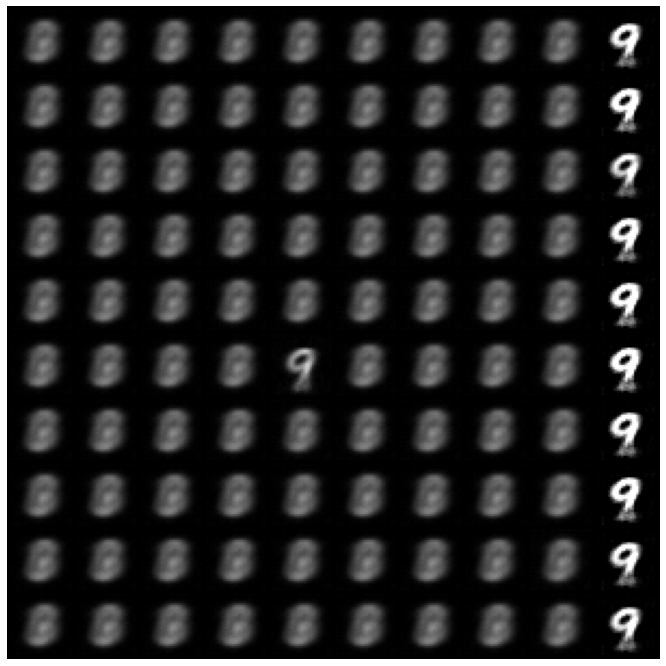}
		%		\caption{}
	\end{subfigure}
	\\
	\begin{subfigure}{0.16\textwidth}
		\includegraphics[height=0.9in]{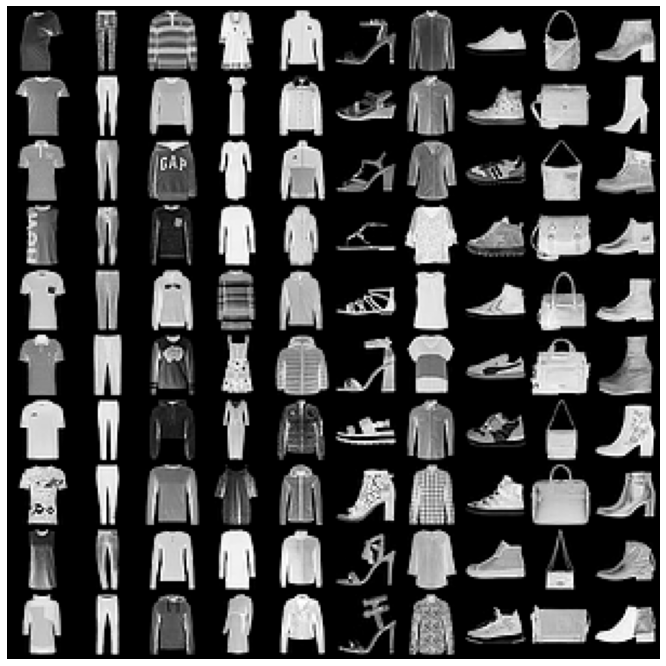}
	\end{subfigure}
	\begin{subfigure}{0.16\textwidth}
		\includegraphics[height=0.9in]{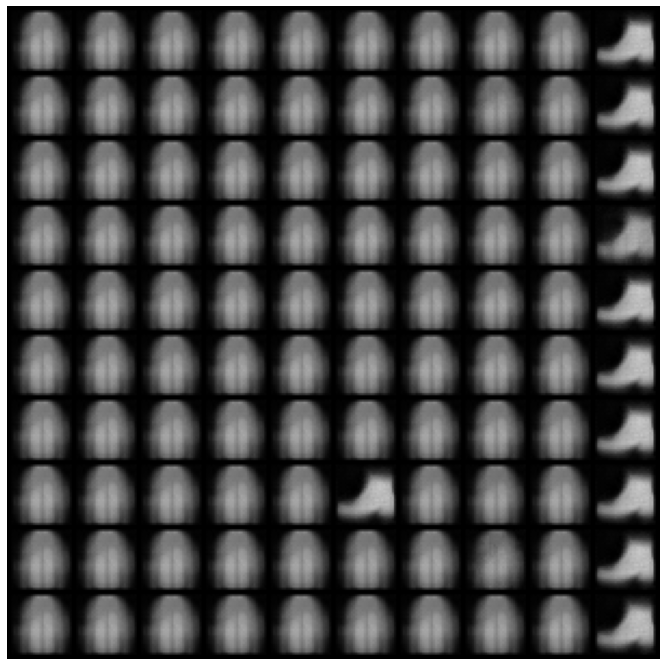}
	\end{subfigure}
	\begin{subfigure}{0.16\textwidth}
		\includegraphics[height=0.9in]{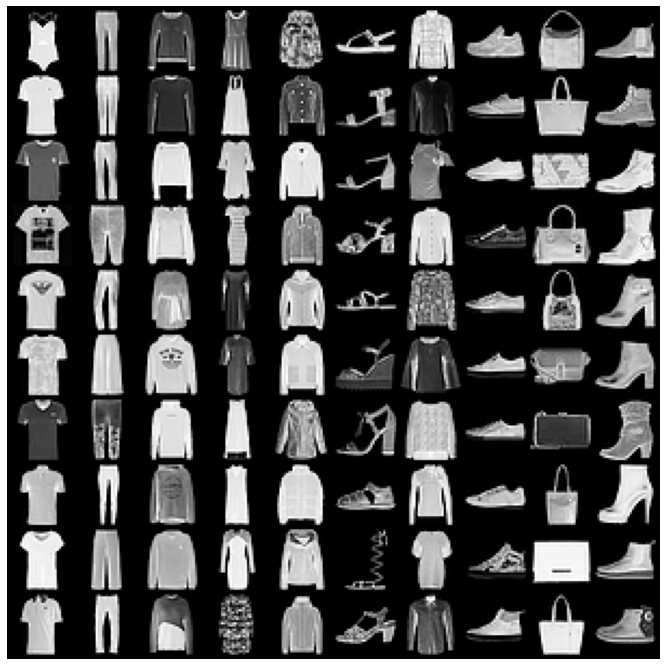}
	\end{subfigure}
	\begin{subfigure}{0.16\textwidth}
		\includegraphics[height=0.9in]{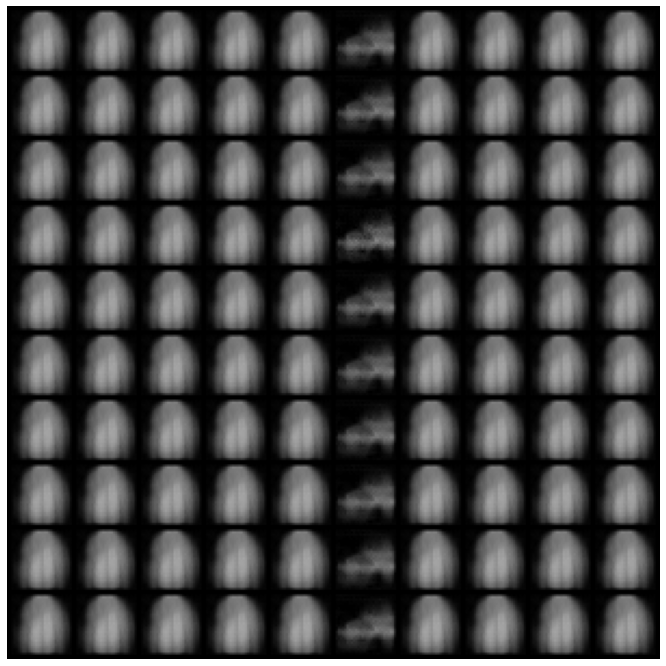}
	\end{subfigure}
	\begin{subfigure}{0.16\textwidth}
		\includegraphics[height=0.9in]{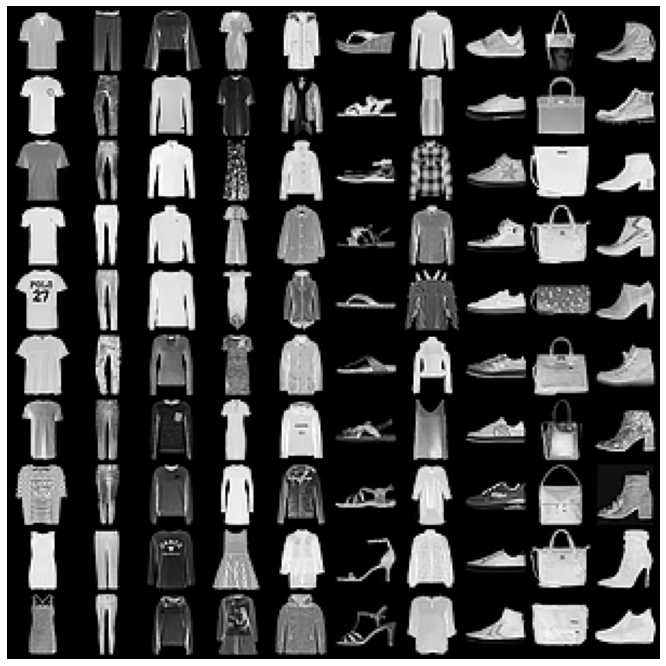}
	\end{subfigure}
	\begin{subfigure}{0.16\textwidth}
		\includegraphics[height=0.9in]{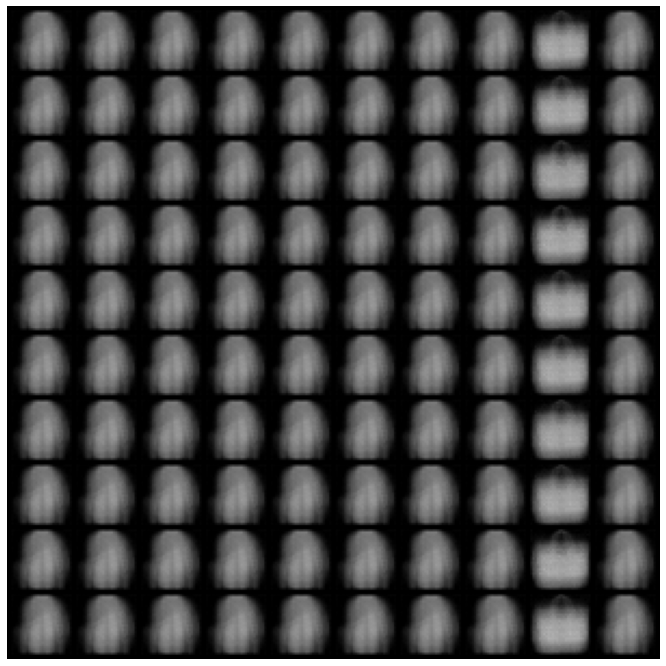}
	\end{subfigure}
	\\
	\begin{subfigure}{0.16\textwidth}
		\includegraphics[height=0.9in]{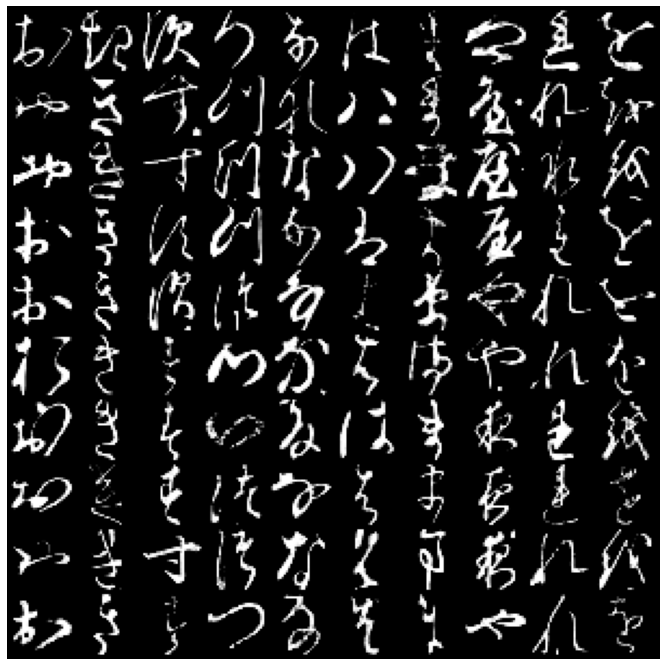}
		%		\caption{}
	\end{subfigure}
	\begin{subfigure}{0.16\textwidth}
		\includegraphics[height=0.9in]{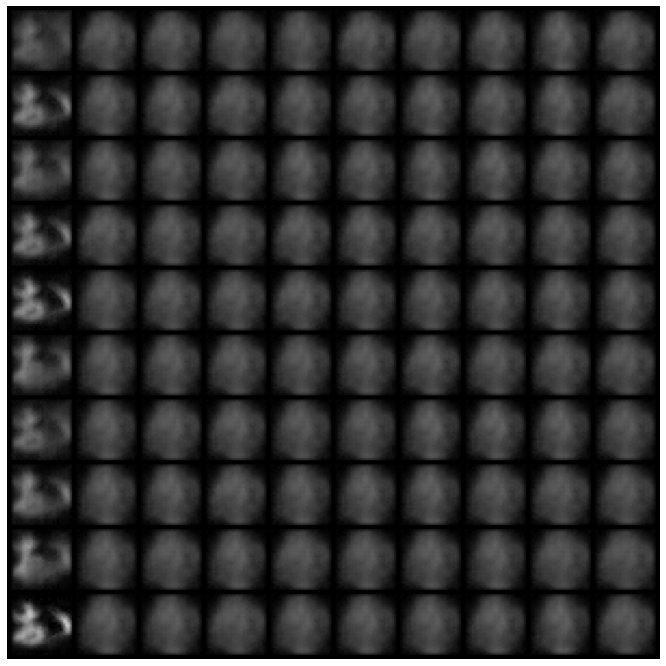}
		%		\caption{}
	\end{subfigure}
	\begin{subfigure}{0.16\textwidth}
		\includegraphics[height=0.9in]{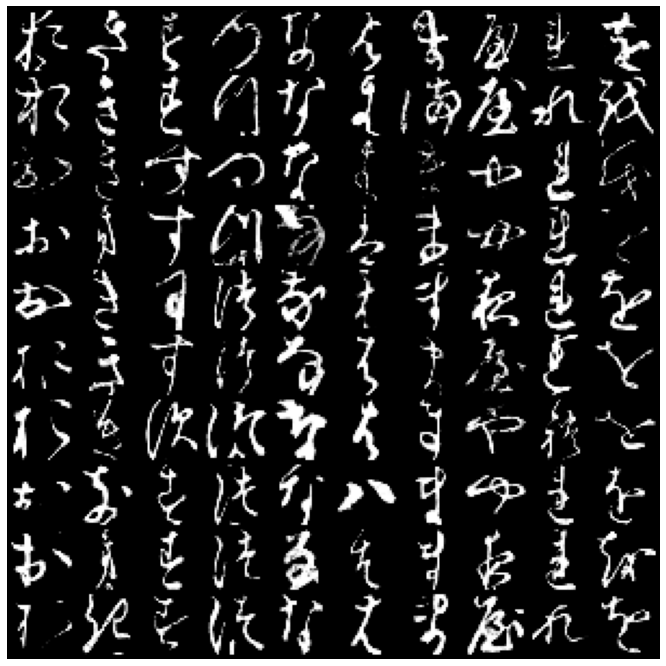}
		%		\caption{}
	\end{subfigure}
	\begin{subfigure}{0.16\textwidth}
		\includegraphics[height=0.9in]{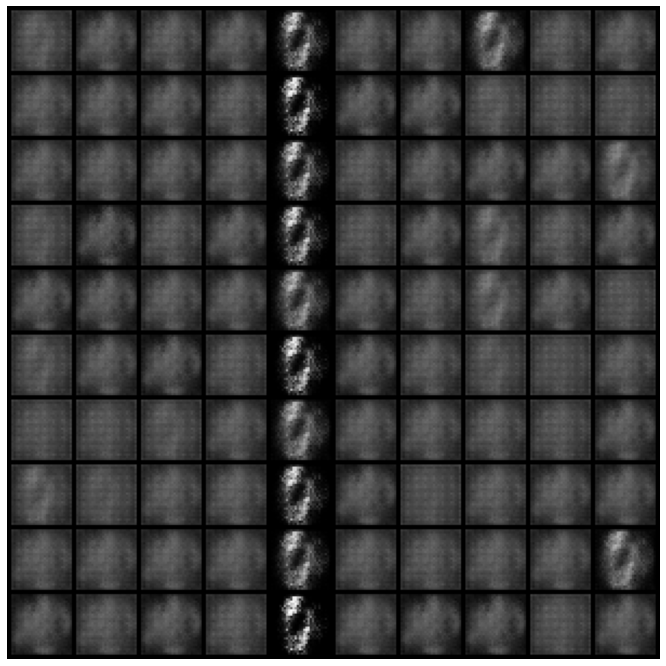}
		%		\caption{}
	\end{subfigure}
	\begin{subfigure}{0.16\textwidth}
		\includegraphics[height=0.9in]{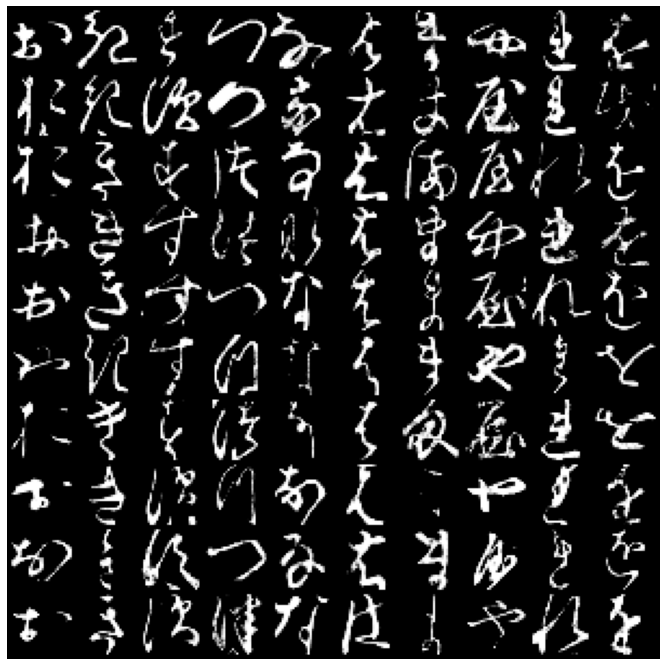}
		%		\caption{}
	\end{subfigure}
	\begin{subfigure}{0.16\textwidth}
		\includegraphics[height=0.9in]{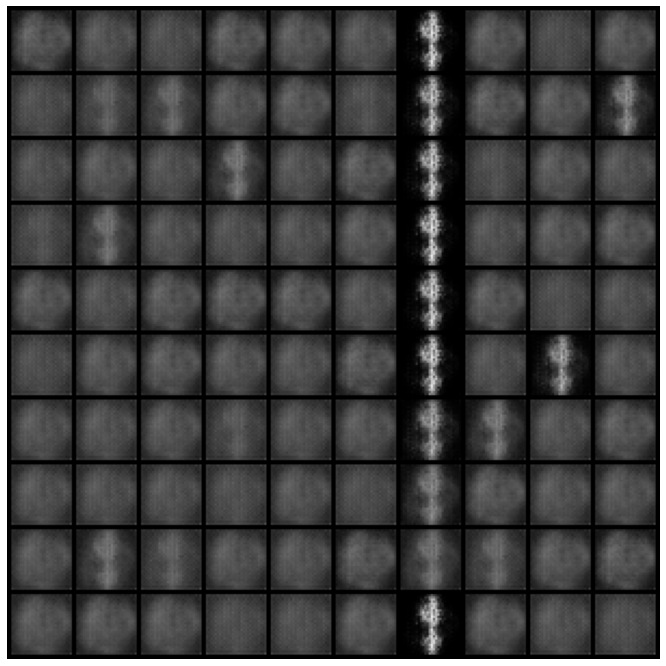}
		%		\caption{}
	\end{subfigure}
	\caption{
		MNIST digits and CausalMIL reconstructions for digits `4', `7', `9'. Figures are best viewed when zoomed. 
		FashionMNIST objects and CausalMIL reconstructions for `dress', `sandal', and `handbag'.	
		More results can be found in the Appendices.
		Figures best viewed when zoomed. }
	\label{Fashion-details}
\end{figure*}

\subsubsection{Qualitative Results}
%Our qualitative results show that CausalMIL effectively learns semantically meaningful representations that are causal to the bag label. 
%Looking at the originals and reconstructions at each row of images in Figure \ref{Fashion-details}, we can see that CausalMIL reconstructs the target digit or fashion objects into their high-level semantic meanings, and such semantics can be viewed as causal to the bag label instead of spurious correlations.

We now qualitatively evaluate the inferred latent factors of CausalMIL by examining the reconstructions of decoder $p_{\bm{f}} (\bm{x}\vert\bm{z})$ using the inferred latent factors in Figure \ref{Fashion-details}. 
For example, in the first row we show the original images and reconstructions of MNIST-bags using `4', `7', and `9' as positive instances.
In the reconstructions of `4' we can see that CausalMIL learns the semantic causal characteristics of the positive instances while ignoring the others. 
The `4's are reconstructed with a `U' shape at the top and a short line at the bottom, while other digits are reconstructed to noise.
The `U' shape of `4' is clear because it is crucial for the positive instance; however, the line at the bottom is blurry since it can be written in different angles and the angles are irrelevant. 
The reconstructions from FashionMNIST-bags and KuzushijiMNIST-bags datasets also corroborate the above results.

\begin{table*}[!t]
	\centering
	\caption{Classification performances on distributionally biased ColoredMNIST and ColoredFashionMNIST datasets in terms of accuracy $\pm$ standard deviation. CausalMIL (ours) is weakly supervised whereas the other approaches are supervised.
%	The means $\pm$ standard deviations are obtained from averaging the results of 10 MNIST/FashionMNIST-bag datasets where each one is repeated for 5 times. 
	}
%	\resizebox{1\linewidth}{!}{
		\begin{tabular}{l | c c| c c}
			\hline
			&	\multicolumn{2}{c|}{ColoredMNIST} & \multicolumn{2}{c}{ColoredFashionMNIST} \\
			\hline
			& Train & Test & Train &  Test\\
			\hline
			ERM & 0.849$\pm$.002 & 0.105$\pm$.007 & 0.832$\pm$.010 & 0.225$\pm$.007  \\
			ERM1 & 0.848$\pm$.002 & 0.109$\pm$.005 & 0.813$\pm$.014 & 0.333$\pm$.089 \\
			ERM2 & 0.850$\pm$.002 & 0.101$\pm$.002 & 0.844$\pm$.019 & 0.132$\pm$.008\\
			ROBUST MIN MAX & 0.843$\pm$.004 & 0.152$\pm$.025 & 0.828$\pm$.001 & 0.292$\pm$.086 \\
			IRM & 0.593$\pm$.044 & 0.628$\pm$.096 & 0.750$\pm$.003 & 0.553$\pm$.124 \\ 
			IRM GAME & 0.634$\pm$.011 & 0.599$\pm$.027 & 0.690$\pm$.101 & 0.702$\pm$.015 \\
			iCaRL & 0.706$\pm$.008  & 0.688$\pm$.007 &  0.750$\pm$.004 &  0.736$\pm$.006\\
%			\hline
			MIVAE & 0.810$\pm$.005 & 0.156$\pm$.003 & 0.823$\pm$.003 & 0.284$\pm$.120\\
			\hline
			\textbf{CausalMIL}& \textbf{0.919$\pm$.002} & \textbf{0.892$\pm$.002} & \textbf{0.873$\pm$.007} & \textbf{0.816$\pm$.015} \\
			\hline
	\end{tabular}
	\label{Quantitative-Colored}
\end{table*}

\subsection{Out-of-Distribution Generalization}
To further validate whether CausalMIL identifies the causal representations $\bm{z}^c$, we evaluate CausalMIL on out-of-distribution (OOD) generalization tasks and compare it supervised algorithms designed specifically for OOD generalization \cite{Arjovsky2019,Ahuja2020,lu2022}. 

% and also a multi-class BiasedMNIST dataset with more extreme distribution biases \cite{}. 
We utilize two widely-adopted biased classification tasks, ColoredMNIST and ColoredFashionMNIST that are commonly used in the invariant risk minimization literature \cite{Arjovsky2019}.
The datasets are constructed following the same procedure as in IRM, where the task is to classify whether a digit is less than 5. The instance labels are first binarized and randomly flipped with 25\% probability. 
Then, the images are colored by green with probabilities $p_e$ which vary across environments, while the rest $1-p_e$ of the images are colored by red. 
There are three environments (two training, one test) where $p_e=0.1, 0.2,$ and $0.9$, respectively. 
Therefore, simply predicting colors instead of digits will lead to high accuracy in the training environments, but the correlation is reversed in the test environment.
For CausalMIL, we construct multi-instance bags based on the flipped instance labels utilizing instances from both training environments. 
Only the bag labels are provided to CausalMIL during training, while the instances from the test environment are used for testing.

We compare the instance label prediction performances of CausalMIL against two categories of \textit{supervised algorithms}. The first category includes standard Empirical Risk Minimization using CNN trained with both (ERM) and each of the training environment (ERM1 and ERM2), and ROBUST MIN MAX which minimizes the maximum loss across different environments. 
The second category are specifically designed OOD generalization algorithms including Invariant Risk Minimization(IRM) \cite{Arjovsky2019}, IRM GAME \cite{Ahuja2020}, and iCaRL \cite{lu2022}. 
Considering that most of the baseline results come from iCaRL \cite{lu2022}, we set CausalMIL to use the same convolutional structure as \cite{lu2022} for fair comparison.
%Existing MIL algorithms are omitted as they are not competitive on the biased test sets.

The results are reported in Table \ref{Quantitative-Colored}. We can see that CausalMIL performs significantly better than all compared supervised methods. 
There are two reasons behind this.
The first reason is that CausalMIL effectively forces the biases into $\bm{z}^e$ by disentangling the causal from non-causal factors, and uses only $\bm{z}^c$ for prediction. 
Although previous supervised OOD generalization algorithms, i.e., IRM and iCaRL, aim to identify and only use the causes of $y$ to improve robustness, they mainly rely on the instance-level labels for identifiability and utilize the scarcely available environments for robustness. Unfortunately, as the instance-level labels are biased from the training environments, their approaches are less effective.
On the contrary, in CausalMIL we rely on the abundance of bag information for identification.
The second reason is that the formulation of MIL is naturally more robust to instance-level label noises as positive bags naturally consist of both positive and negative instances. Furthermore, previous theoretical results have also shown that generative MIL algorithm \cite{Doran2016} is effective even when negative bags are falsely labeled as positive. 
%as previous theoretical results \cite{Doran2016} have shown that learning instance concepts in MIL can be reduced to learning with semi-random one-sided label noise, which itself is PAC-learnable. 
%However, we point out that negative bags may also contain positive instances (flipped into negative) in the biased datasets \cite{Arjovsky2019}, which could worth further investigation since no previous MIL algorithms have reported empirical results regarding noisy negative bags.

%\subsection{Discussion}
%There are two reasons behind the effectiveness of CausalMIL.
%On the one hand, the representations learned by previous MIL algorithms do not attempt to capture the underlying semantic information that is causal to the bag label.
%From this perspective, CausalMIL can be viewed as possibly the first attempt towards learning causal representations by utilizing the problem characteristics of MIL.
%On the other hand, most of the existing algorithms treat the inexact bag label solely as the source of ambiguity that prohibits accurate prediction;
%however, finding some way to utilize this ambiguity is crucial for MIL performances. 
%Although there are some previous efforts \cite{Zhou2009,Zhang2020a,Rymarczyk2021}, CausalMIL is the first to utilize the bag information for learning meaningful representations.
%Therefore, CausalMIL is a synergistic framework that utilizes the problem characteristics of MIL for learning better latent representations, which in turn benefits instance label prediction performances.
%

\section{Related Work}
Most MIL algorithms can be categorized into two groups according to whether they work at the bag space or at the instance space.
Bag space algorithms \cite{Zhou2009,Wei2017,Zhang2020a,Feng2021} work by embedding multi-instance bags into a single feature vector representation and then solving the single-instance learning problem in the embedded space, and therefore are capable of predicting instance labels.
Instance space MIL algorithms aim to directly separate the positive instances from the negative ones \cite{Andrews2002,Kandemir2014,Haussmann2017,Wang2021}. 
These algorithms can be used to predict instance labels, although not all of them are explicitly designed for the task.However, these algorithms only work with pre-computed features.

Recently, several deep learning-based MIL algorithms have been proposed by utilizing permutation-invariant pooling operations, and can be used for instance label prediction.
For example, \cite{Wang2018} used a max-pooling layer with a fully connected neural network; \cite{Ilse2018} introduced the attention mechanism as a permutation-invariant MIL pooling operation and used the attention weights for instance label prediction;
\cite{Rymarczyk2021} proposed to integrate self-attention with the attention mechanism for capturing the non-i.i.d. information among instances; \cite{Li2021a} utilized Gumbel reparametrization and proposed an algorithm for the generalized MIL assumption.
Unfortunately, none of the above methods learn semantically meaningful representations. 

Unsupervised learning of meaningful representations with VAE-based models has attracted much attention \cite{Higgins2017,Chen2018}. Since it has been shown that unsupervised disentanglement is theoretically impossible without inductive biases \cite{Locatello2019}, various methods have been proposed for learning disentangled representations using different forms of weak supervisions.
One form of weakly-supervised disentanglement learning that is closely related to MIL is group-based VAE \cite{Bouchacourt2018,Hosoya2019}.
Similar to MIL where instances are organized into bags, group-based algorithms require the objects to be divided into groups:
those within the same group share the same \textit{content} but have different \textit{styles} that are independent of the contents, while those among different groups should have different contents.

Recently, MIVAE \cite{Zhang2021} is proposed based on connections between MIL and group-based disentangled representation learning which utilizes VAE for capturing the dependencies among instances as the shared content factors. However, MIVAE also does not learn semantically meaningful latents, nor does it provide identifiability on what latents it actually learns because of its unconditional prior. Please refer to the Appendices for a more detailed comparison between CausalMIL and MIVE. 

\section{Conclusion}
In this work, we proposed CausalMIL, an algorithm that learns semantically meaningful causal representations from multi-instance bags and the corresponding bag-level supervision. The learned representations not only significantly improve the performances of traditional MIL tasks such as instance label prediction, but also exhibits notable performances in supervised learning tasks such as OOD generalization. Qualitative and quantative evaluation results show that CausalMIL performs significantly better than existing state-of-the-art deep MIL algorithms on semi-synthetic and real-world datasets.

There are some limitations of our method and many future directions worth investigating. For example, CausalMIL is only designed for the standard multi-instance assumption. It would be interesting to see if the results extend to other assumptions in multi-instance learning. A possible way for such extension is to utilize the Gumbel reparameterization as discussed in \cite{Li2021a}. Another possible direction would be developing an extension to learning from positive unlabeled (PU) examples, another weakly supervised learning problem that is closely related to MIL.

\begin{ack}
The authors wish to thank the anonymous reviewers for their constructive comments and suggestions.
This work was supported by the National Science Foundation of China (62176055, 62206047, 72204110).
We thank the Big Data Center of Southeast University for providing the facility support on the numerical calculations in this paper.
\end{ack}

\bibliographystyle{plain}
\bibliography{reference}

\begin{thebibliography}{10}

\bibitem{Ahuja2020}
Kartik Ahuja, Karthikeyan Shanmugam, Kush Varshney, and Amit Dhurandhar.
\newblock Invariant risk minimization games.
\newblock In {\em Proceedings of the 37th International Conference on Machine
  Learning}, PMLR 119, pages 145--155, 2020.

\bibitem{Andrews2002}
Stuart Andrews, Ioannis Tsochantaridis, and Thomas Hofmann.
\newblock Support vector machines for multiple-instance learning.
\newblock In {\em Advances in Neural Information Processing Systems 15}, pages
  577--584, 2002.

\bibitem{Angelidis2018}
Stefanos Angelidis and Mirella Lapata.
\newblock Multiple instance learning networks for fine-grained sentiment
  analysis.
\newblock {\em Transactions of the Association for Computational Linguistics},
  (6):17--34, 2018.

\bibitem{Arjovsky2019}
Martin Arjovsky, Léon Bottou, Ishaan Gulrajani, and David Lopez-Paz.
\newblock Invariant risk minimization.
\newblock {\em arXiv preprint}, arXiv: 1907.02893, 2019.

\bibitem{Bouchacourt2018}
Diane Bouchacourt, Ryota Tomioka, and Sebastian Nowozin.
\newblock Multi-level variational autoencoder: Learning disentangled
  representations from grouped observations.
\newblock In {\em Proceedings of the 32nd AAAI Conference on Artificial
  Intelligence}, pages 2095--2102, 2018.

\bibitem{Chen2018}
Ricky T.~Q. Chen, Xuechen Li, Roger Grosse, and David Duvenaud.
\newblock Isolating sources of disentanglement in variational autoencoders.
\newblock In {\em Advances in Neural Information Processing Systems 31}, page
  2615–2625, 2018.

\bibitem{Clanuwat2018}
Tarin Clanuwat, Mikel Bober-Irizar, Asanobu Kitamoto, Alex Lamb, Kazuaki
  Yamamoto, and David Ha.
\newblock Deep learning for classical {Japanese} literature.
\newblock {\em arXiv preprint}, arXiv:1812.01718, 2018.

\bibitem{Dietterich1997}
Thomas~G. Dietterich, Richard~H. Lathrop, and Tom{\'{a}}s Lozano-P{\'{e}}rez.
\newblock Solving the multiple instance problem with axis-parallel rectangles.
\newblock {\em Artificial Intelligence}, 89(1-2):31--71, 1997.

\bibitem{Doran2016}
Gary Doran and Soumya Ray.
\newblock Multiple-instance learning from distributions.
\newblock {\em Journal of Machine Learning Research}, 17(128):1--50, 2016.

\bibitem{Feng2021}
Lei Feng, Senlin Shu, Yuzhou Cao, Lue Tao, Hongxin Wei, Tao Xiang, Bo~An, and
  Gang Niu.
\newblock Multiple-instance learning from similar and dissimilar bags.
\newblock In {\em Proceedings of the 27th ACM SIGKDD Conference on Knowledge
  Discovery and Data Mining}, page 374–382, 2021.

\bibitem{Foulds2010}
James Foulds and Eibe Frank.
\newblock A review of multi-instance learning assumptions.
\newblock {\em The Knowledge Engineering Review}, 25(1):1--25, 2010.

\bibitem{Haussmann2017}
Manuel Haussmann, Fred~A. Hamprecht, and Melih Kandemir.
\newblock Variational bayesian multiple instance learning with gaussian
  processes.
\newblock In {\em 2017 {IEEE} Conference on Computer Vision and Pattern
  Recognition}, pages 6570--6579, 2017.

\bibitem{Higgins2017}
Irina Higgins, Loic Matthey, Arka Pal, Christopher Burgess, Xavier Glorot,
  Matthew Botvinick, Shakir Mohamed, and Alexander Lerchner.
\newblock {beta-VAE}: Learning basic visual concepts with a constrained
  variational framework.
\newblock In {\em Proceedings of the 5th International Conference on Learning
  Representations}. https://openreview.net/forum?id=Sy2fzU9gl, 2017.

\bibitem{Hosoya2019}
Haruo Hosoya.
\newblock Group-based learning of disentangled representationswith
  generalizability for novel contents.
\newblock In {\em Proceedings of the 28th International Joint Conference on
  Artificial Intelligence}, pages 2506--2513, 2019.

\bibitem{Ilse2018}
Maximilian Ilse, Jakub Tomczak, and Max Welling.
\newblock Attention-based deep multiple instance learning.
\newblock In {\em Proceedings of the 35th International Conference on Machine
  Learning}, PMLR 80, pages 2127--2136, 2018.

\bibitem{Kandemir2014}
Melih Kandemir and Fred~A. Hamprecht.
\newblock Instance label prediction by dirichlet process multiple instance
  learning.
\newblock In {\em Proceedings of the 30th Conference on Uncertainty in
  Artificial Intelligence}, pages 380--389, 2014.

\bibitem{Khemakhem2020}
Ilyes Khemakhem, Diederik~P. Kingma, Ricardo~Pio Monti, and Aapo Hyv{\"a}rinen.
\newblock Variational autoencoders and nonlinear {ICA}: A unifying framework.
\newblock In {\em Proceedings of the 23rd International Conference on
  Artificial Intelligence and Statistcs}, PMLR 108, pages 2007--2017, 2020.

\bibitem{Kingma2014}
Diederik~P Kingma and Max Welling.
\newblock Auto-encoding variational bayes.
\newblock In {\em Proceedings of the 2nd International Conference on Learning
  Representations}. https://openreview.net/forum?id=33X9fd2-9FyZd, 2014.

\bibitem{Lecun1998}
Y.~Lecun, L.~Bottou, Y.~Bengio, and P.~Haffner.
\newblock Gradient-based learning applied to document recognition.
\newblock {\em Proceedings of the {IEEE}}, 86(11):2278--2324, 1998.

\bibitem{Li2021}
Bin Li, Yin Li, and Kevin~W Eliceiri.
\newblock Dual-stream multiple instance learning network for whole slide image
  classification with self-supervised contrastive learning.
\newblock In {\em Proceedings of the IEEE/CVF Conference on Computer Vision and
  Pattern Recognition}, pages 14318--14328, 2021.

\bibitem{Li2021a}
Xin-Chun Li, De-Chuan Zhan, Jia-Qi Yang, and Yi~Shi.
\newblock Deep multiple instance selection.
\newblock {\em Science China Information Sciences}, 64(3), 2021.

\bibitem{Locatello2019}
Francesco Locatello, Stefan Bauer, Mario Lucic, Gunnar Raetsch, Sylvain Gelly,
  Bernhard Schölkopf, and Olivier Bachem.
\newblock Challenging common assumptions in the unsupervised learning of
  disentangled representations.
\newblock In {\em Proceedings of the 36th International Conference on Machine
  Learning}, PMLR 97, pages 4114--4124, 2019.

\bibitem{lu2022}
Chaochao Lu, Yuhuai Wu, Jos{\'e}~Miguel Hern{\'a}ndez-Lobato, and Bernhard
  Sch{\"o}lkopf.
\newblock Invariant causal representation learning for out-of-distribution
  generalization.
\newblock In {\em International Conference on Learning Representations}.
  https://openreview.net/forum?id=-e4EXDWXnSn, 2022.

\bibitem{Lu2021}
Ming~Y. Lu, Tiffany~Y. Chen, Drew~F.K. Williamson, Melissa Zhao, Maha Shady,
  Jana Lipkova, and Faisal Mahmood.
\newblock {AI}-based pathology predicts origins for cancers of unknown primary.
\newblock {\em Nature}, 594:106--110, 6 2021.

\bibitem{Mita2021}
Graziano Mita, Maurizio Filippone, and Pietro Michiardi.
\newblock An identifiable double vae for disentangled representations.
\newblock In {\em Proceedings of the 38th International Conference on Machine
  Learning, PMLR 139}, pages 7769--7779, 2021.

\bibitem{Pearl2009}
Judea Pearl.
\newblock {\em Causality}.
\newblock Cambridge University Press, 2009.

\bibitem{Peters2016}
Jonas Peters, Peter Bühlmann, and Nicolai Meinshausen.
\newblock Causal inference by using invariant prediction: identification and
  confidence intervals.
\newblock {\em Journal of the Royal Statistical Society: Series B (Statistical
  Methodology)}, 78(5):947--1012, 2016.

\bibitem{Jan2000}
Jan Ramon and Luc~De Raedt.
\newblock Multi-instance neural networks.
\newblock In {\em Proceedings of the ICML-2000 Workshop on Attribute-value and
  Relational Learning}, pages 53--60, 2000.

\bibitem{Rymarczyk2021}
Dawid Rymarczyk, Adriana Borowa, Jacek Tabor, and Bartosz Zielinsk.
\newblock Kernel self-attention for weakly-supervised image classification
  using deepmultiple instance learning.
\newblock In {\em Proceedings of the IEEE/CVF Winter Conference on Applications
  of Computer Vision}, pages 1721--1730, 2021.

\bibitem{Scholkopf2021}
Bernhard Sch{\"o}lkopf, Francesco Locatello, Stefan Bauer, Nan~Rosemary Ke, Nal
  Kalchbrenner, Anirudh Goyal, and Yoshua Bengio.
\newblock Toward causal representation learning.
\newblock {\em Proceedings of the {IEEE}}, 109(5):612--634, 2021.

\bibitem{Shi2020}
Xiaoshuang Shi, Fuyong Xing, Yuanpu Xie, Zizhao Zhang, Lei Cui, and Lin Yang.
\newblock Loss-based attention for deep multiple instance learning.
\newblock {\em Proceedings of the 34th {AAAI} Conference on Artificial
  Intelligence}, pages 5742--5749, 2020.

\bibitem{Sirinukunwattana2016}
Korsuk Sirinukunwattana, Shan E~Ahmed Raza, Yee-Wah Tsang, David R.~J. Snead,
  Ian~A. Cree, and Nasir~M. Rajpoot.
\newblock Locality sensitive deep learning for detection and classification of
  nuclei in routine colon cancer histology images.
\newblock {\em {IEEE} Transactions on Medical Imaging}, 35(5):1196--1206, 2016.

\bibitem{Spirtes2000}
Peter Spirtes, Clark Glymour, and Richard Scheines.
\newblock {\em Causation, Prediction, and Search}.
\newblock The MIT Press, 2000.

\bibitem{Wang2021}
Fulton Wang and Ali Pinar.
\newblock The multiple instance learning gaussian process probit model.
\newblock In {\em Proceedings of The 24th International Conference on
  Artificial Intelligence and Statistics, PMLR 130}, pages 3034--3042, 2021.

\bibitem{Wang2018}
Xinggang Wang, Yongluan Yan, Peng Tang, Xiang Bai, and Wenyu Liu.
\newblock Revisiting multiple instance neural networks.
\newblock {\em Pattern Recognition}, 74:15--24, 2018.

\bibitem{Wang2019}
Yun Wang, Juncheng Li, and Florian Metze.
\newblock A comparison of five multiple instance learning pooling functions for
  sound event detection with weak labeling.
\newblock In {\em 2019 IEEE International Conference on Acoustics, Speech and
  Signal Processing (ICASSP)}, pages 31--35, 2019.

\bibitem{Wei2017}
Xiu-Shen Wei, Jianxin Wu, and Zhi-Hua Zhou.
\newblock Scalable algorithms for multi-instance learning.
\newblock {\em {IEEE} Transactions on Neural Networks and Learning Systems},
  28(4):975--987, 2017.

\bibitem{Xiao2017}
Han Xiao, Kashif Rasul, and Roland Vollgraf.
\newblock {Fashion-MNIST}: a novel image dataset for benchmarking machine
  learning algorithms.
\newblock {\em arXiv preprint}, arXiv:1708.07747, 2017.

\bibitem{Zaheer2017}
Manzil Zaheer, Satwik Kottur, Siamak Ravanbakhsh, Barnabas Poczos, Russ~R
  Salakhutdinov, and Alexander~J Smola.
\newblock Deep sets.
\newblock In {\em Advances in Neural Information Processing Systems 30}, 2017.

\bibitem{Zhang2021}
Weijia Zhang.
\newblock Non-i.i.d. multi-instance learning for predicting instance and bag
  labels using variational autoencoder.
\newblock In {\em Proceedings of the 30th International Joint Conference on
  Artificial Intelligence}, pages 3377--3383, 2021.

\bibitem{Zhang2020a}
Weijia Zhang, Lin Liu, and Jiuyong Li.
\newblock Robust multi-instance learning with stable instances.
\newblock In {\em Proceedings of the 24th European Conference on Artificial
  Intelligence}, pages 1682--1689, 2020.

\bibitem{Zhou2017}
Zhi-Hua Zhou.
\newblock A brief introduction to weakly supervised learning.
\newblock {\em National Science Review}, 5(1):44--53, 2017.

\bibitem{Zhou2009}
Zhi-Hua Zhou, Yu-Yin Sun, and Yu-Feng Li.
\newblock Multi-instance learning by treating instances as non-i.i.d. samples.
\newblock In {\em Proceedings of the 26th International Conference on Machine
  Learning}, pages 1249--1256, 2009.

\end{thebibliography}

%%%%%%%%%%%%%%%%%%%%%%%%%%%%%%%%%%%%%%%%%%%%%%%%%%%%%%%%%%%%
\section*{Checklist}
%%%% BEGIN INSTRUCTIONS %%%
%The checklist follows the references.  Please
%read the checklist guidelines carefully for information on how to answer these
%questions.  For each question, change the default \answerTODO{} to \answerYes{},
%\answerNo{}, or \answerNA{}.  You are strongly encouraged to include a {\bf
%justification to your answer}, either by referencing the appropriate section of
%your paper or providing a brief inline description.  For example:
%\begin{itemize}
%  \item Did you include the license to the code and datasets? \answerYes{See Section~\ref{gen_inst}.}
%  \item Did you include the license to the code and datasets? \answerNo{The code and the data are proprietary.}
%  \item Did you include the license to the code and datasets? \answerNA{}
%\end{itemize}
%%%% END INSTRUCTIONS %%%
\begin{enumerate}
\item For all authors...
\begin{enumerate}
  \item Do the main claims made in the abstract and introduction accurately reflect the paper's contributions and scope?
    \answerYes{}
  \item Did you describe the limitations of your work?
    \answerYes{}
  \item Did you discuss any potential negative societal impacts of your work?
    \answerNA{}
  \item Have you read the ethics review guidelines and ensured that your paper conforms to them?
    \answerNA{}
\end{enumerate}
\item If you are including theoretical results...
\begin{enumerate}
  \item Did you state the full set of assumptions of all theoretical results?
    \answerYes{}
        \item Did you include complete proofs of all theoretical results?
    \answerYes{In the Appendix.}
\end{enumerate}
\item If you ran experiments...
\begin{enumerate}
  \item Did you include the code, data, and instructions needed to reproduce the main experimental results (either in the supplemental material or as a URL)?
    \answerYes{}
  \item Did you specify all the training details (e.g., data splits, hyperparameters, how they were chosen)?
    \answerYes{}
        \item Did you report error bars (e.g., with respect to the random seed after running experiments multiple times)?
    \answerYes{}
        \item Did you include the total amount of compute and the type of resources used (e.g., type of GPUs, internal cluster, or cloud provider)?
    \answerYes{}
\end{enumerate}
\item If you are using existing assets (e.g., code, data, models) or curating/releasing new assets...
\begin{enumerate}
  \item If your work uses existing assets, did you cite the creators?
    \answerYes{}
  \item Did you mention the license of the assets?
    \answerNA{}
  \item Did you include any new assets either in the supplemental material or as a URL?
    \answerNA{}
  \item Did you discuss whether and how consent was obtained from people whose data you're using/curating?
    \answerNA{}
  \item Did you discuss whether the data you are using/curating contains personally identifiable information or offensive content?
    \answerNA{}
\end{enumerate}
\item If you used crowdsourcing or conducted research with human subjects...
\begin{enumerate}
  \item Did you include the full text of instructions given to participants and screenshots, if applicable?
    \answerNA{}
  \item Did you describe any potential participant risks, with links to Institutional Review Board (IRB) approvals, if applicable?
    \answerNA{}
  \item Did you include the estimated hourly wage paid to participants and the total amount spent on participant compensation?
    \answerNA{}
\end{enumerate}
\end{enumerate}

%%%%%%%%%%%%%%%%%%%%%%%%%%%%%%%%%%%%%%%%%%%%%%%%%%%%%%%%%%%%

\clearpage
\appendix

\setcounter{figure}{0}
\setcounter{table}{0}
\section{Appendices}
\subsection{Additional Experiment Results}
\begin{figure*}[!h]
	\centering
	\begin{subfigure}{0.22\textwidth}
		\includegraphics[height=1.2in]{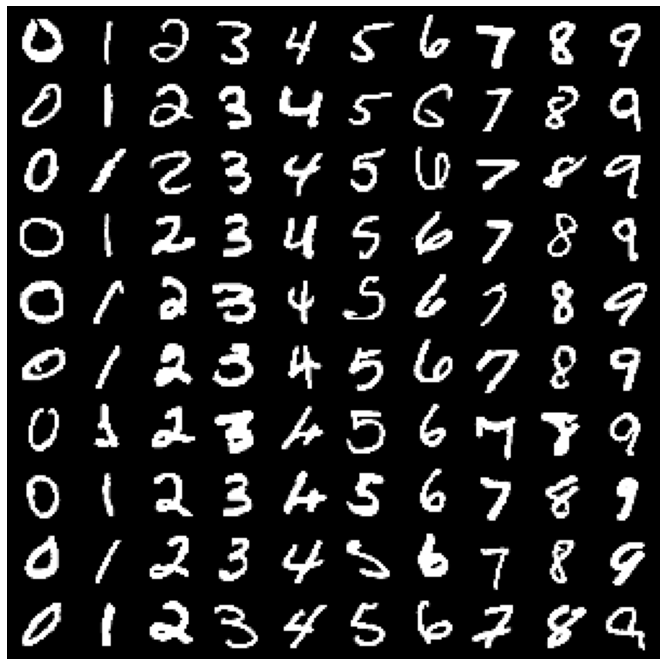}
	\end{subfigure}
	\begin{subfigure}{0.22\textwidth}
		\includegraphics[height=1.2in]{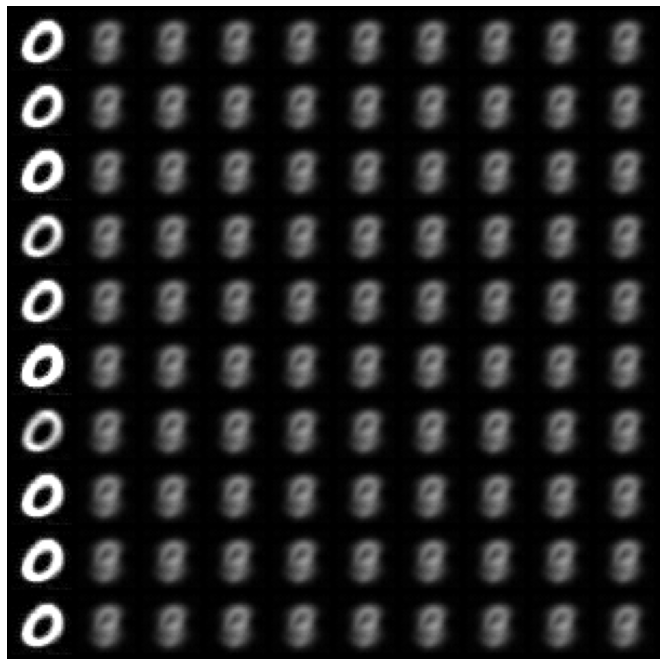}
	\end{subfigure}
	\begin{subfigure}{0.22\textwidth}
		\includegraphics[height=1.2in]{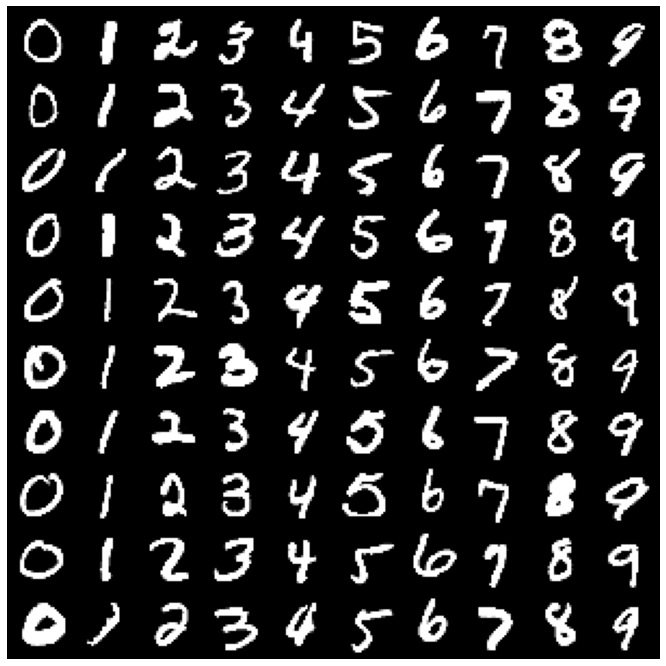}
	\end{subfigure}
	\begin{subfigure}{0.22\textwidth}
		\includegraphics[height=1.2in]{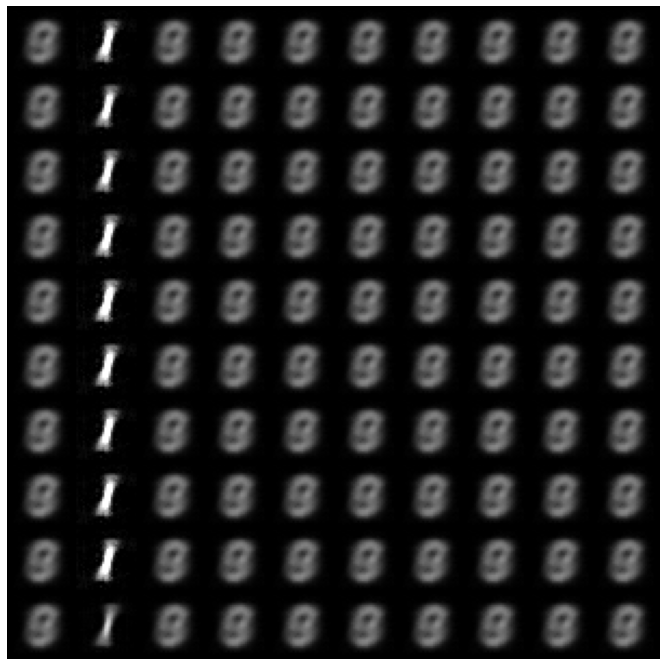}
	\end{subfigure}
	
	\begin{subfigure}{0.22\textwidth}
		\includegraphics[height=1.2in]{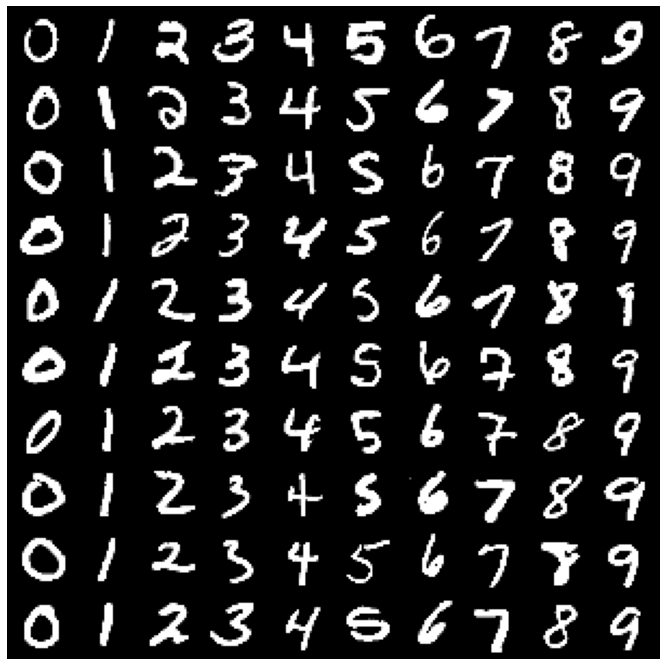}
	\end{subfigure}
	\begin{subfigure}{0.22\textwidth}
		\includegraphics[height=1.2in]{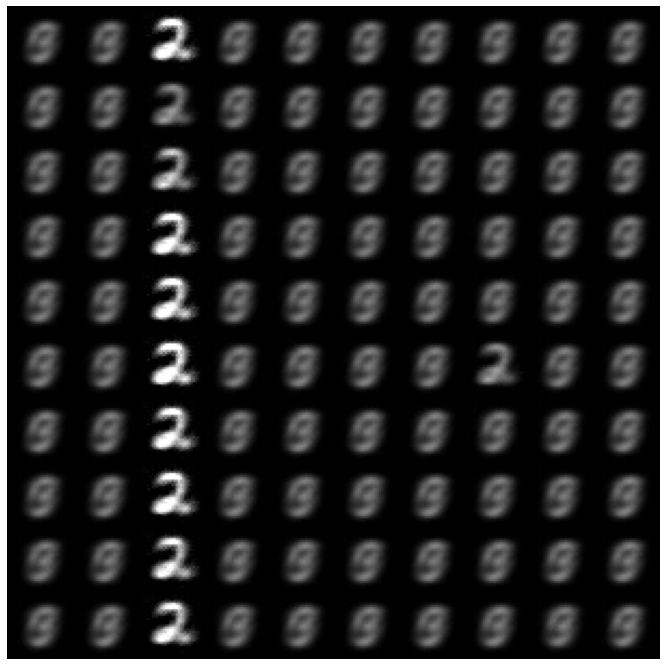}
	\end{subfigure}
	\begin{subfigure}{0.22\textwidth}
		\includegraphics[height=1.2in]{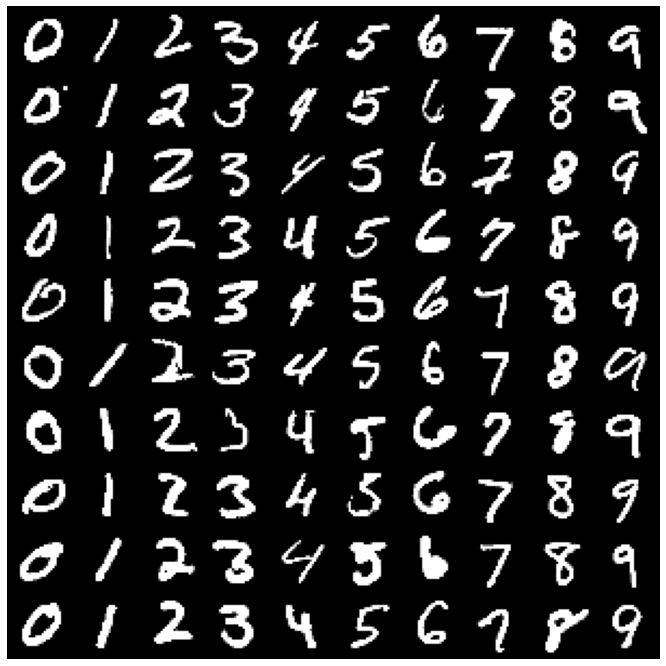}
	\end{subfigure}
	\begin{subfigure}{0.22\textwidth}
		\includegraphics[height=1.2in]{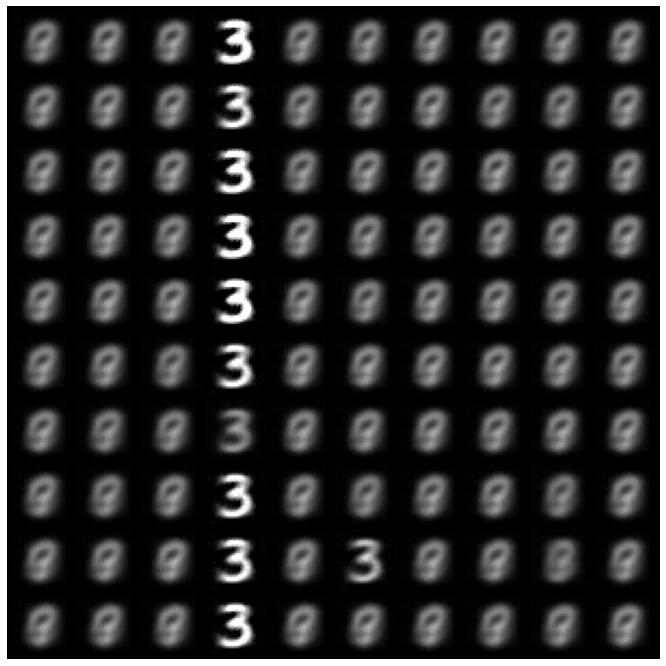}
	\end{subfigure}
	
	\begin{subfigure}{0.22\textwidth}
		\includegraphics[height=1.2in]{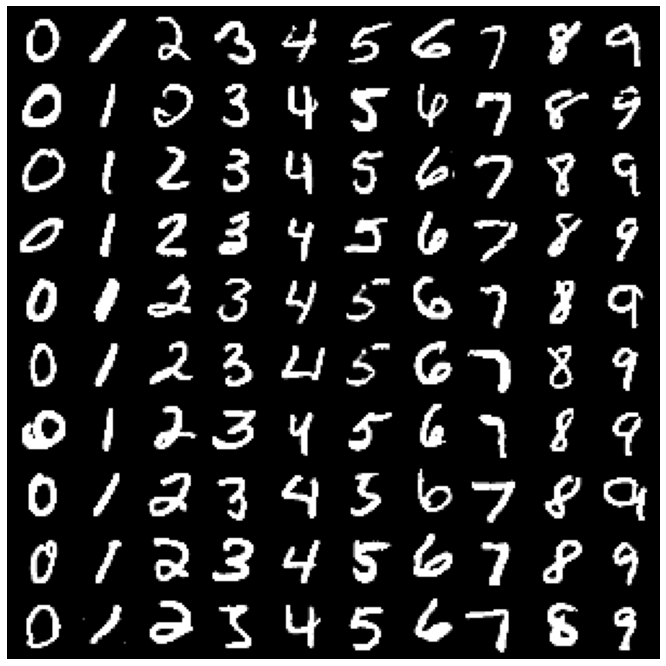}
	\end{subfigure}
	\begin{subfigure}{0.22\textwidth}
		\includegraphics[height=1.2in]{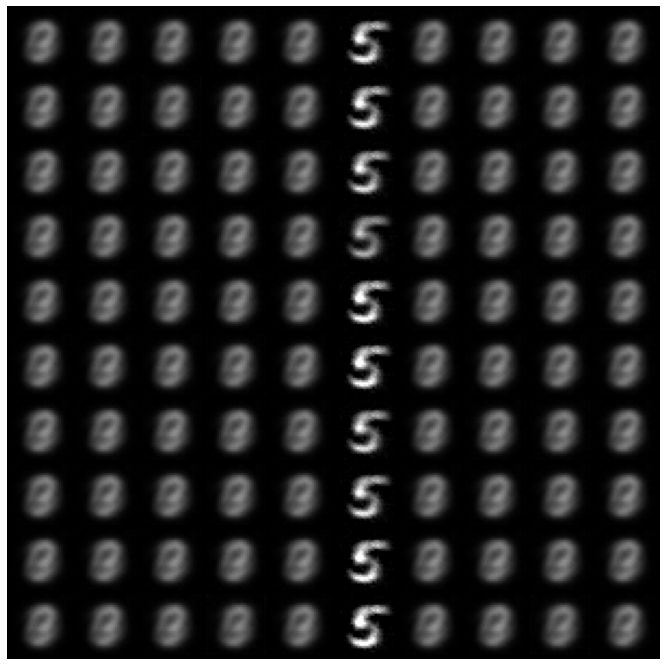}
	\end{subfigure}
	\begin{subfigure}{0.22\textwidth}
		\includegraphics[height=1.2in]{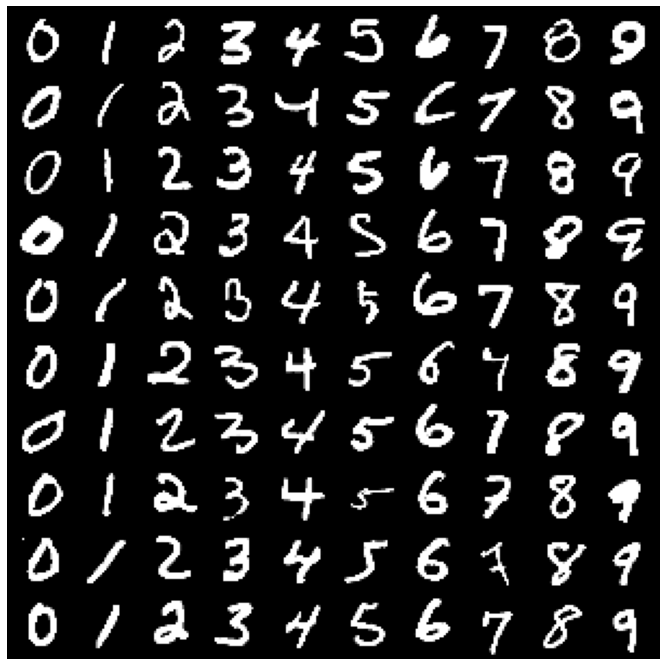}
	\end{subfigure}
	\begin{subfigure}{0.22\textwidth}
		\includegraphics[height=1.2in]{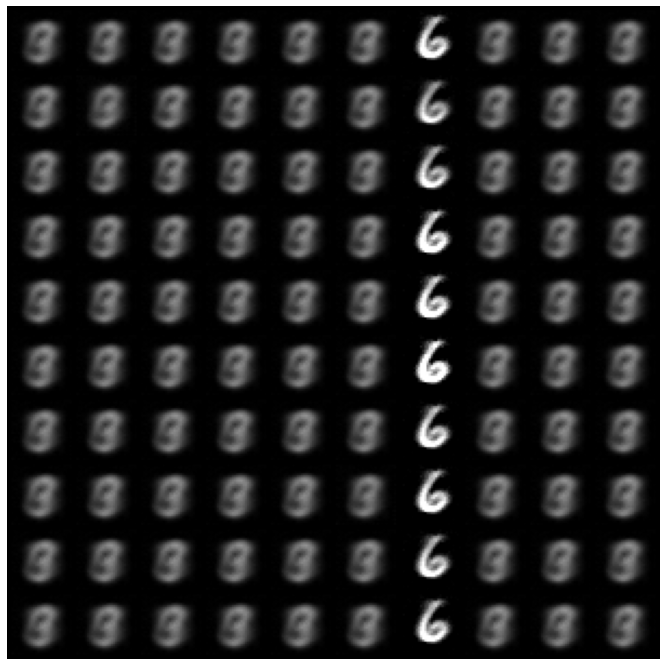}
	\end{subfigure}
	
	\begin{subfigure}{0.22\textwidth}
		\includegraphics[height=1.2in]{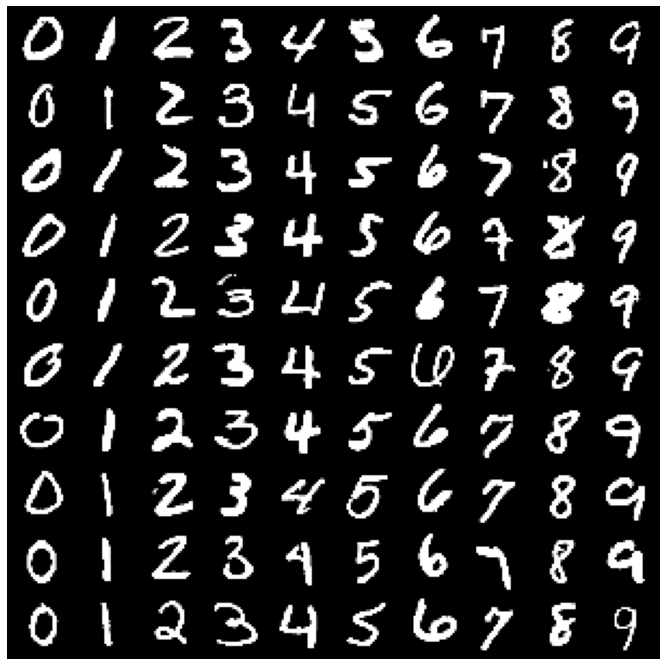}
	\end{subfigure}
	\begin{subfigure}{0.22\textwidth}
		\includegraphics[height=1.2in]{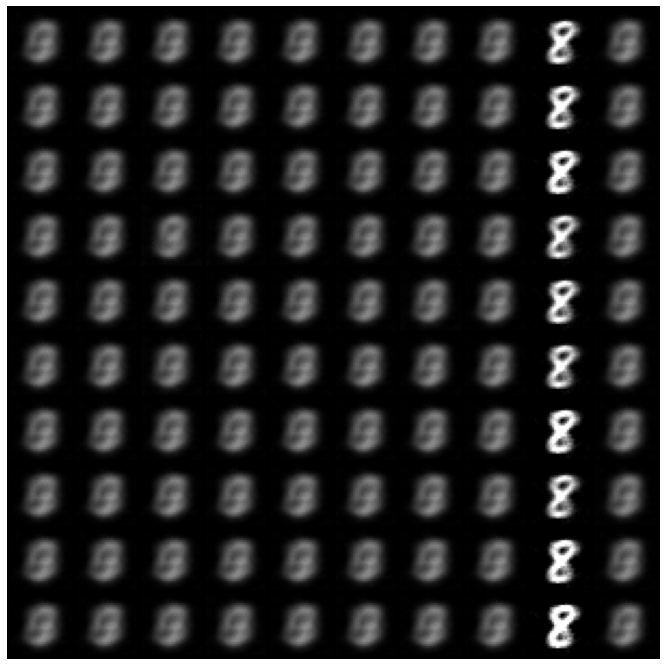}
	\end{subfigure}
	\caption{(a) Original MNIST digits and CausalMIL reconstructions  for digits `0', `1', `2', `3', `5', `6', and `8'. The original digits are randomly sampled from the test set. Figures are best viewed when zoomed.}
	\label{MNIST-details}
\end{figure*}

We now report the qualitative results on MNIST, FashionMNIST, and KuzushijiMNIST datasets that are omitted from the main manuscript. 
We can see that on MNIST-bags, CausalMIL successfully learns semantically meaningful representations for all the digits. 
On FashionMNIST-bags, TargetedMI makes some mistakes among the `pullover', `coat', and `shirt' objects (the 3rd, 5th, and 7th columns). 
This is perhaps because that these objects themselves are difficult to distinguish even for humans, and the original FashionMNIST labels are not noise free. 
For example, looking at the subfigure at the 1st row, 1st column, we can see that some objects belonging to the `shirt' class are not different from those in the `t-shirt' class (for example, the object at the 3rd row, 7th column). Furthermore, it is also difficult to differentiate between `pullover' and `shirt' (the 3rd and the 7th column of each subfigure). 

An interesting observation from the KuzushijiMNIST-bags is that CausalMIL is able to learn the causal invariant representation for hiragana characters that have more than one handwritten forms.
Let us look at the bottom subfigure which depicts 10 hiragana characters and their handwriting. We can see that some of the hiragana  character have two types of handwritten forms. For example, the second hiragana character has two forms from the bottom subfigure; accordingly, in its original vs reconstructions comparisons (first two subfigures of the first row), CausalMIL is successful in recognizing these two different handwritten forms.
%Looking at the second KMNIST hiragana form character and its handwritten forms (the second row in the bottom subfigure of Figure 3), we can see that one hiragana form character has more than one handwritten variations. 
%From the third and fourth subfigures in the first row of Figure 3, it can be seen that CausalMIL successfully learns at least two forms of the target handwriting.

\begin{figure*}[!t]
	\centering
	\begin{subfigure}{0.22\textwidth}
		\includegraphics[height=1.2in]{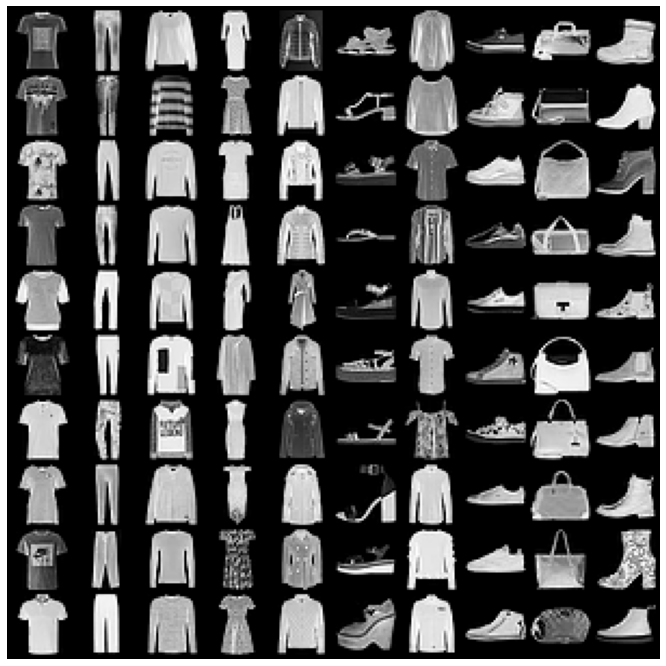}
	\end{subfigure}
	\begin{subfigure}{0.22\textwidth}
		\includegraphics[height=1.2in]{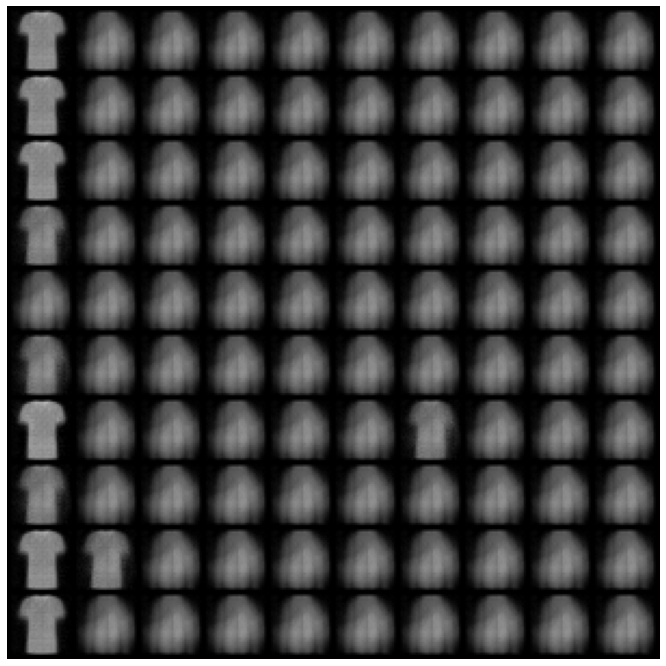}
	\end{subfigure}
	\begin{subfigure}{0.22\textwidth}
		\includegraphics[height=1.2in]{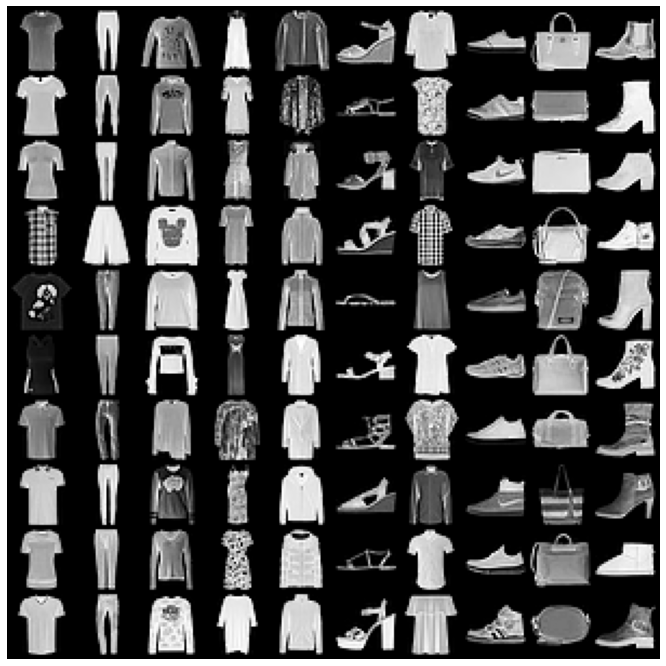}
	\end{subfigure}
	\begin{subfigure}{0.22\textwidth}
		\includegraphics[height=1.2in]{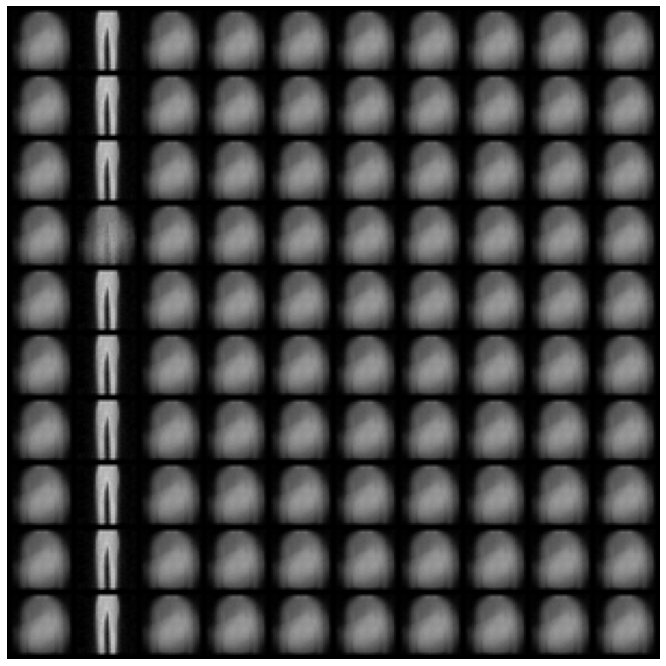}
	\end{subfigure}
	\begin{subfigure}{0.22\textwidth}
		\includegraphics[height=1.2in]{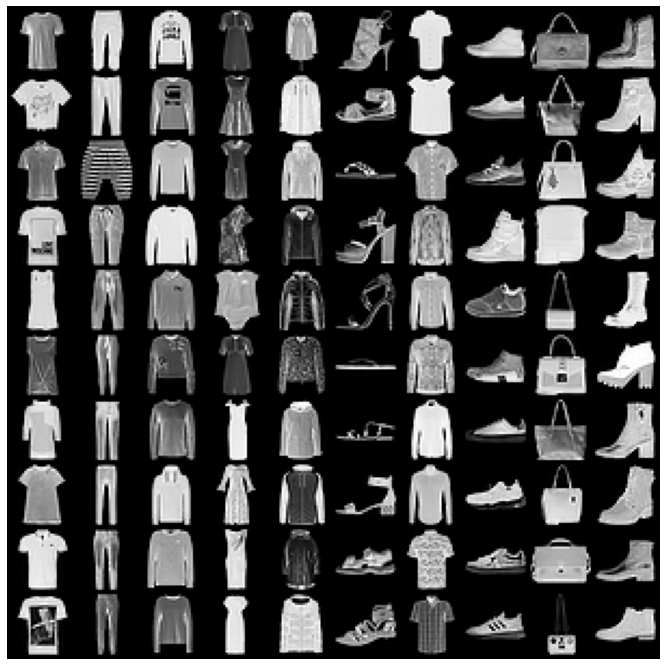}
	\end{subfigure}
	\begin{subfigure}{0.22\textwidth}
		\includegraphics[height=1.2in]{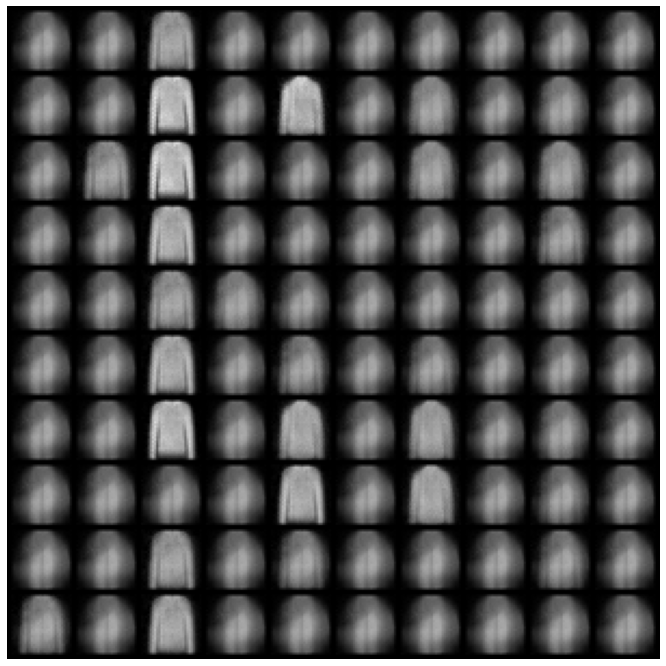}
	\end{subfigure}
	\begin{subfigure}{0.22\textwidth}
		\includegraphics[height=1.2in]{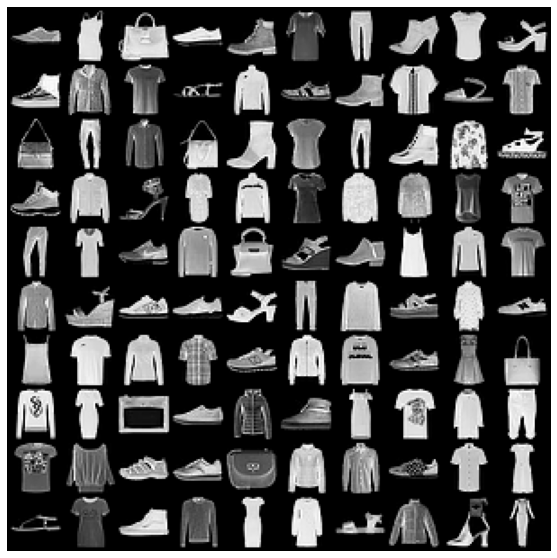}
	\end{subfigure}
	\begin{subfigure}{0.22\textwidth}
		\includegraphics[height=1.2in]{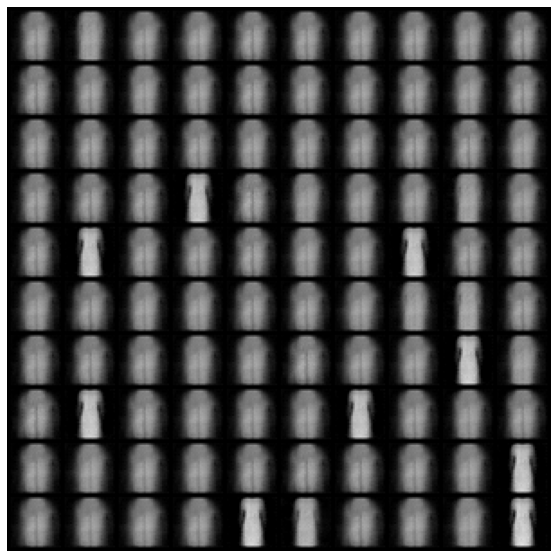}
	\end{subfigure}
	\begin{subfigure}{0.22\textwidth}
		\includegraphics[height=1.2in]{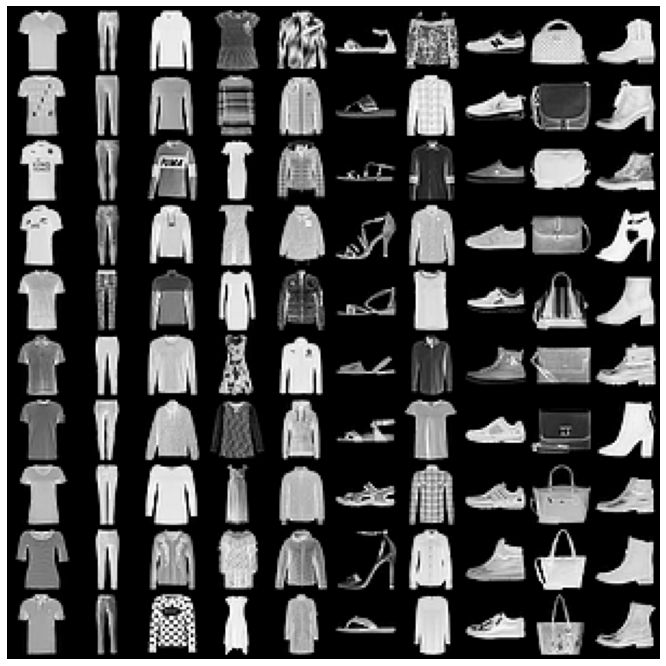}
	\end{subfigure}
	\begin{subfigure}{0.22\textwidth}
		\includegraphics[height=1.2in]{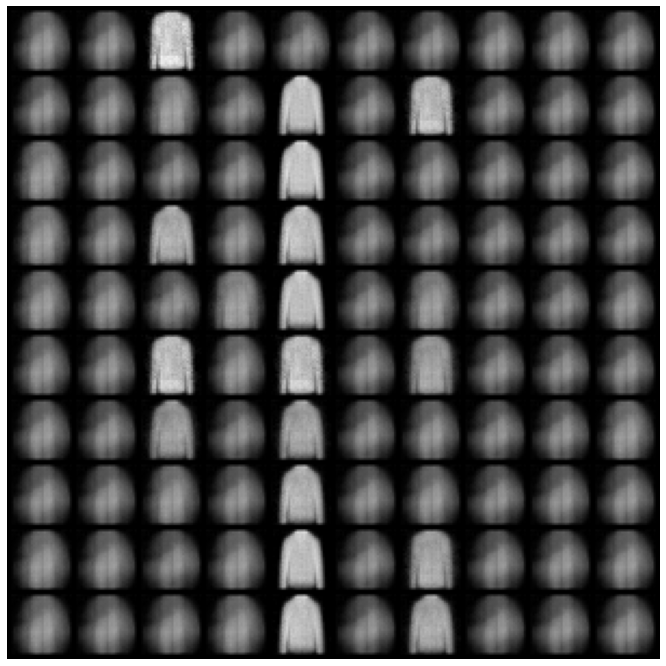}
	\end{subfigure}
	\begin{subfigure}{0.22\textwidth}
		\includegraphics[height=1.2in]{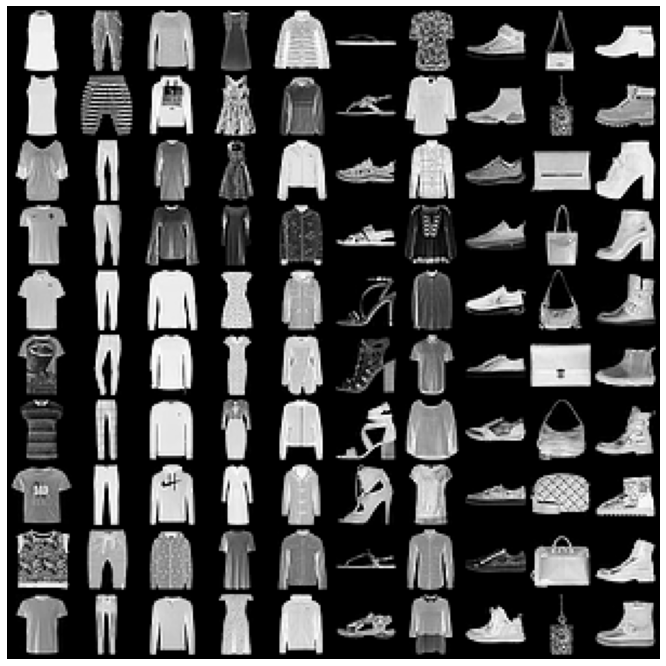}
	\end{subfigure}
	\begin{subfigure}{0.22\textwidth}
		\includegraphics[height=1.2in]{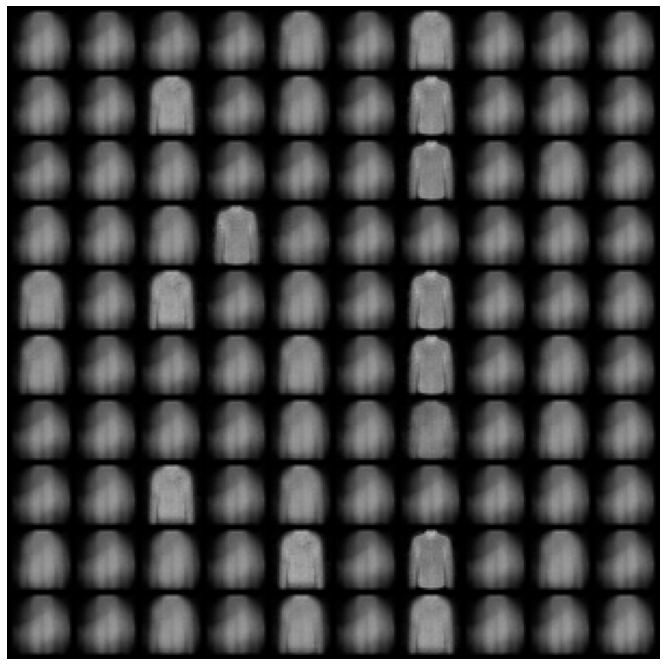}
	\end{subfigure}
	\begin{subfigure}{0.22\textwidth}
		\includegraphics[height=1.2in]{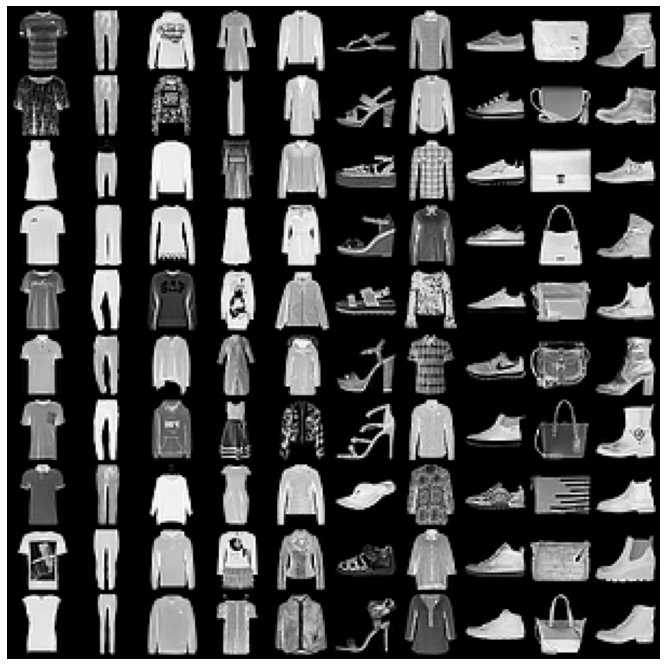}
	\end{subfigure}
	\begin{subfigure}{0.22\textwidth}
		\includegraphics[height=1.2in]{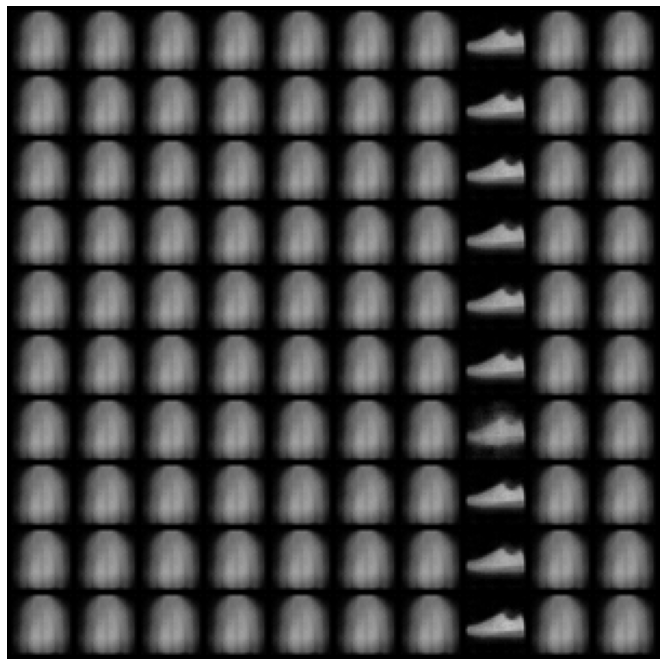}
	\end{subfigure}
	\caption{(a) Original FashionMNIST images and CausalMIL reconstructions for object classes `t-shirt', `trousers', `pullover', `coat', `shirt', `sneaker', and `ankle boot'. The original images are randomly sampled from the test set. Figures are best viewed when zoomed.}
	\label{MNIST-details}
\end{figure*}

\begin{figure*}[!t]
	\centering
	\begin{subfigure}{0.22\textwidth}
			\includegraphics[height=1.2in]{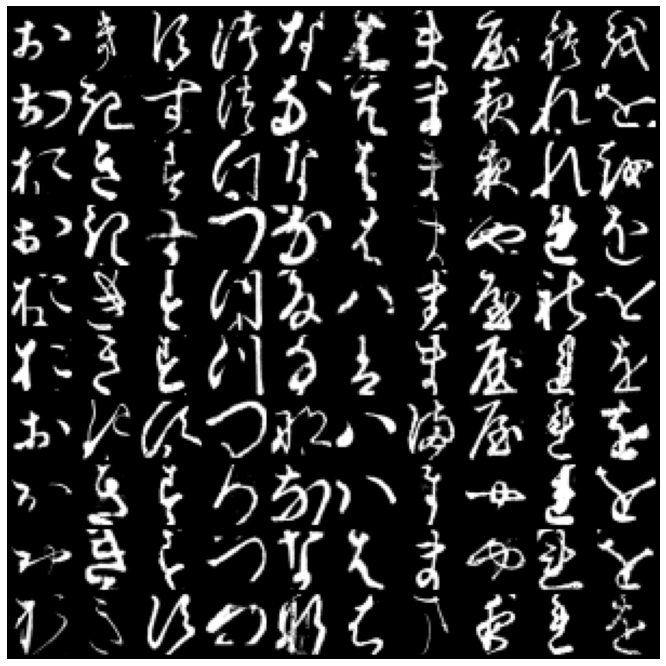}
		\end{subfigure}
	\begin{subfigure}{0.22\textwidth}
			\includegraphics[height=1.2in]{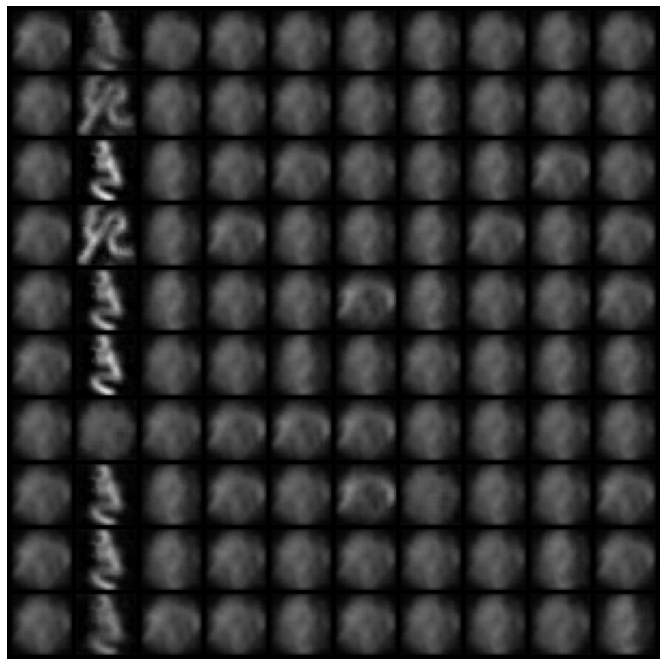}
		\end{subfigure}
	\begin{subfigure}{0.22\textwidth}
			\includegraphics[height=1.2in]{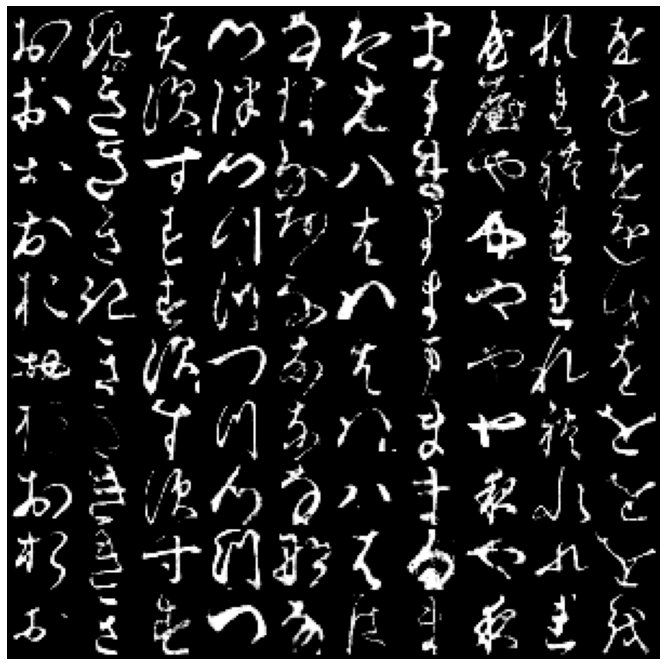}
		\end{subfigure}
	\begin{subfigure}{0.22\textwidth}
			\includegraphics[height=1.2in]{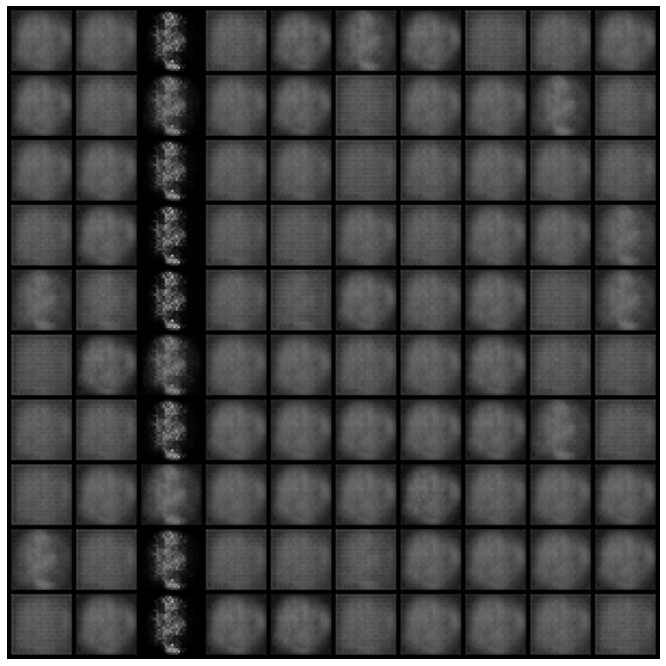}
		\end{subfigure}
	
	\begin{subfigure}{0.22\textwidth}
	\includegraphics[height=1.2in]{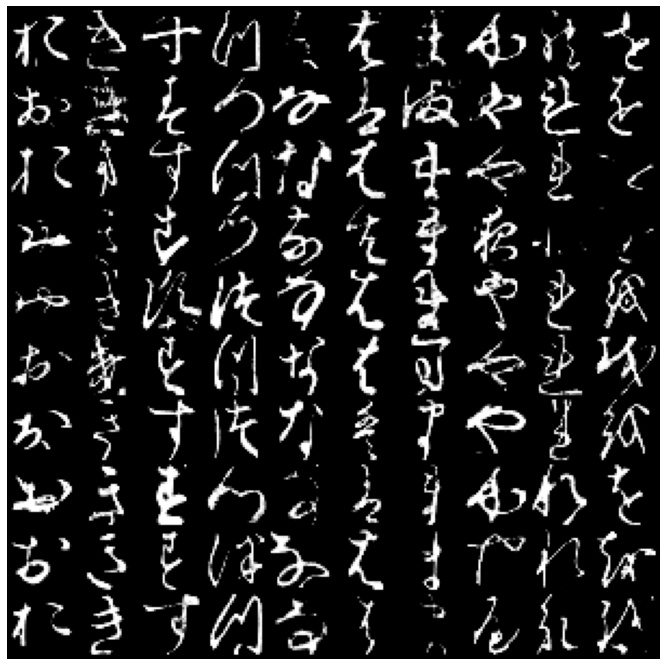}
	\end{subfigure}
	\begin{subfigure}{0.22\textwidth}
		\includegraphics[height=1.2in]{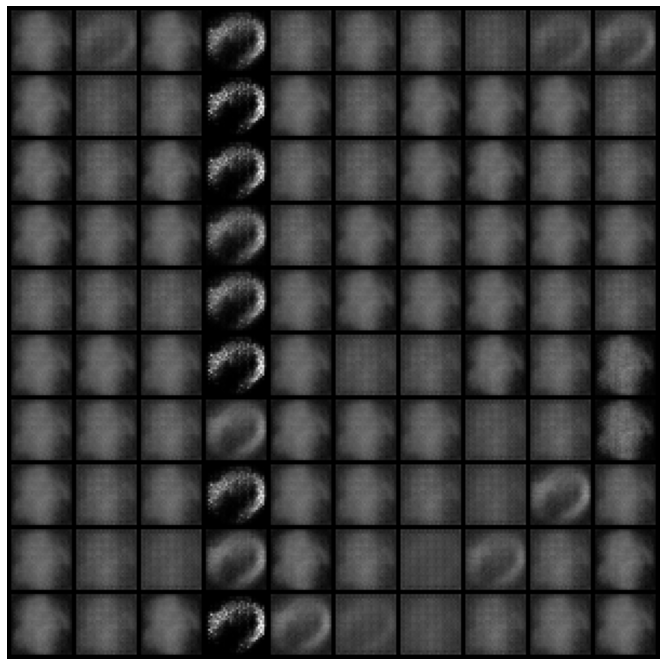}
	\end{subfigure}
	\begin{subfigure}{0.22\textwidth}
		\includegraphics[height=1.2in]{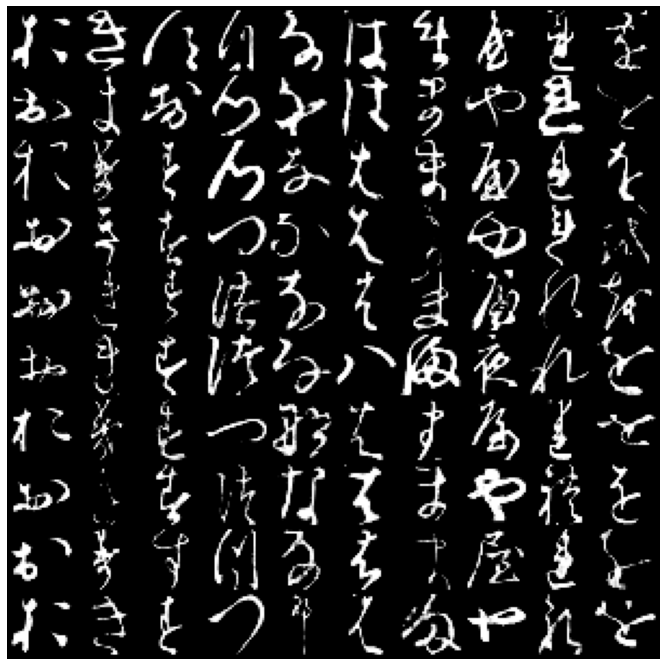}
	\end{subfigure}
	\begin{subfigure}{0.22\textwidth}
		\includegraphics[height=1.2in]{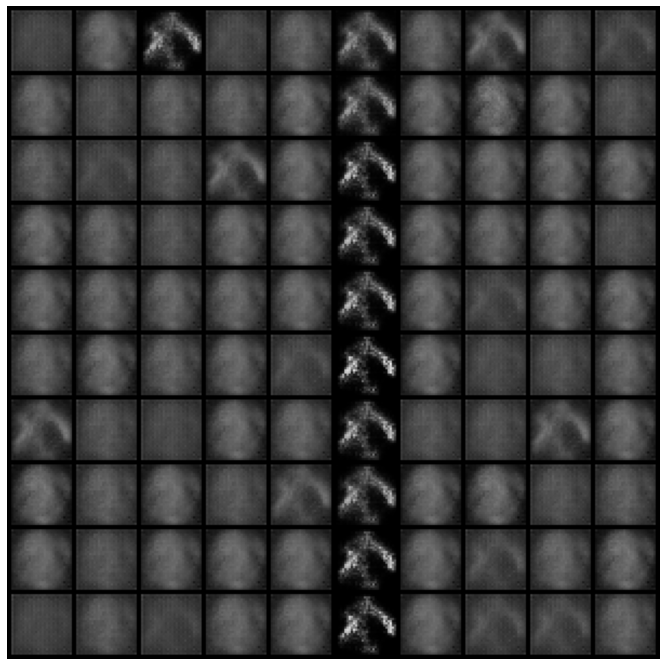}
	\end{subfigure}

	\begin{subfigure}{0.22\textwidth}
	\includegraphics[height=1.2in]{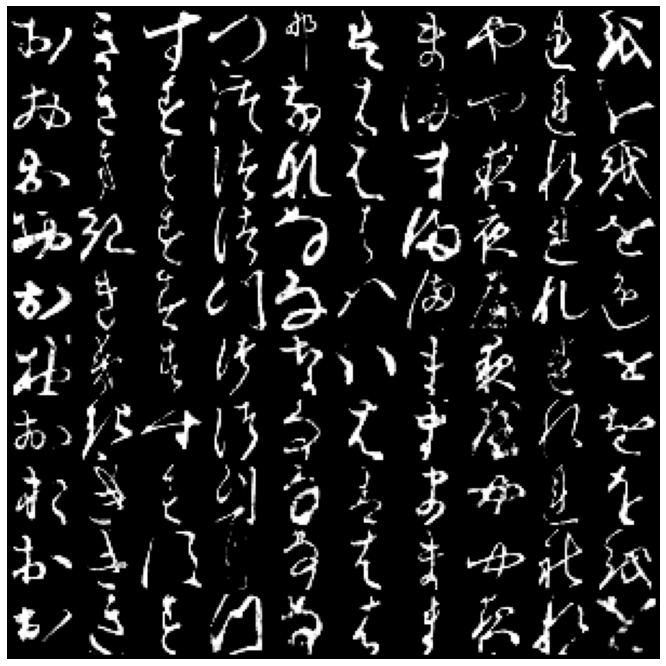}
	\end{subfigure}
	\begin{subfigure}{0.22\textwidth}
		\includegraphics[height=1.2in]{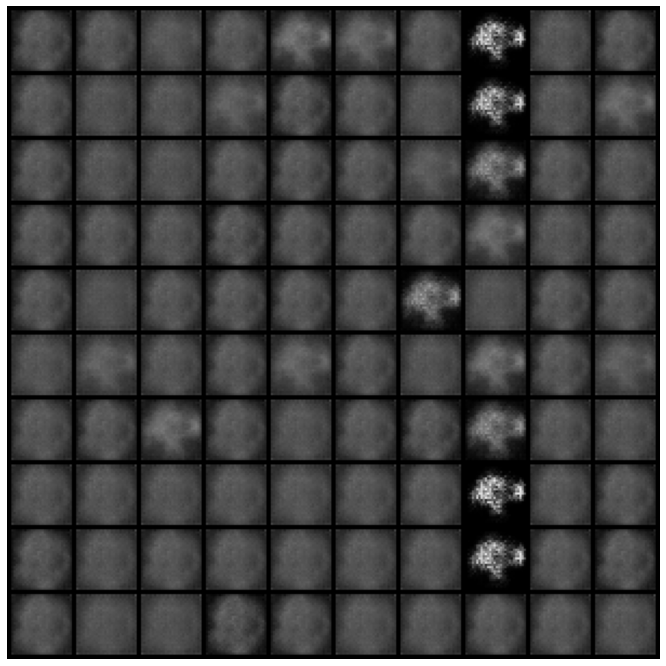}
	\end{subfigure}
	\begin{subfigure}{0.22\textwidth}
		\includegraphics[height=1.2in]{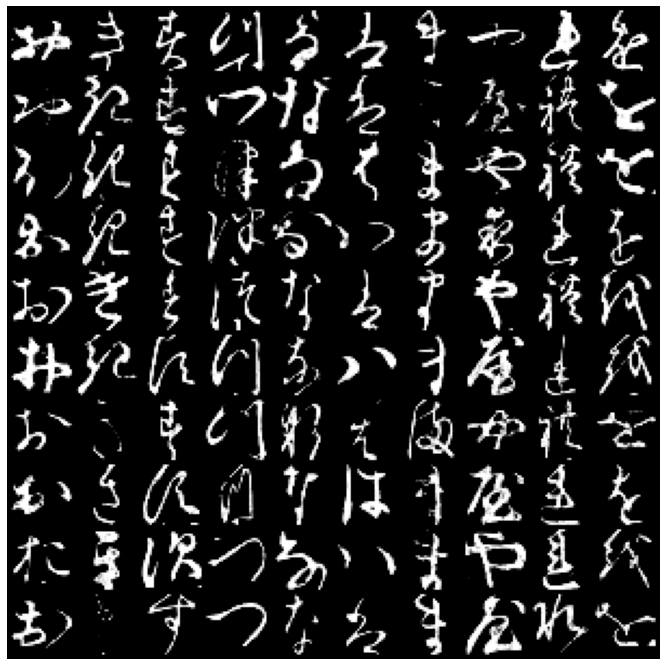}
	\end{subfigure}
	\begin{subfigure}{0.22\textwidth}
		\includegraphics[height=1.2in]{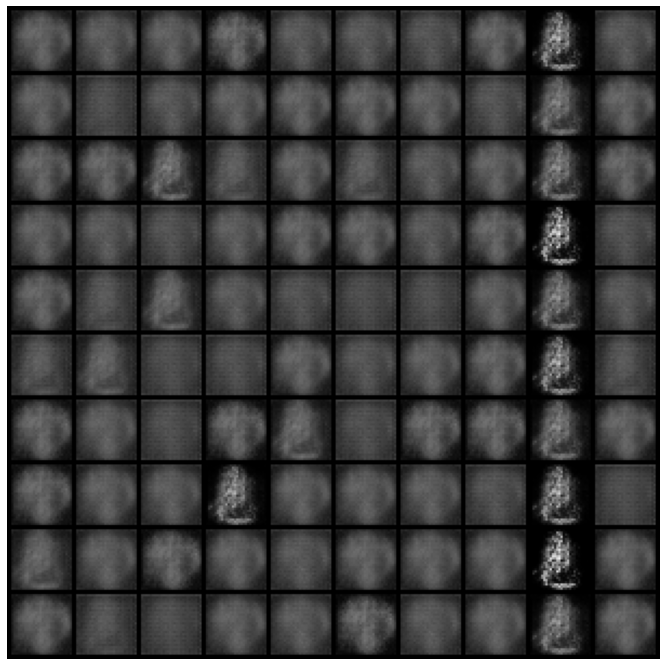}
	\end{subfigure}

	\begin{subfigure}{0.22\textwidth}
	\includegraphics[height=1.2in]{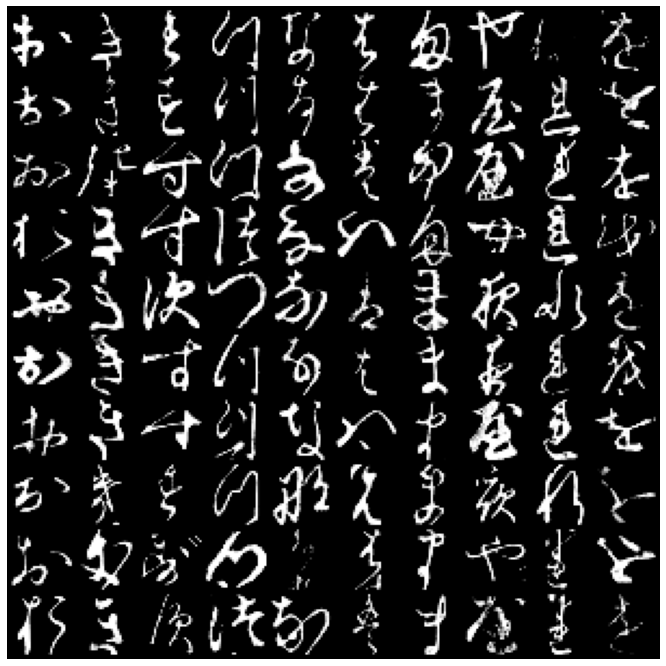}
	\end{subfigure}
	\begin{subfigure}{0.22\textwidth}
		\includegraphics[height=1.2in]{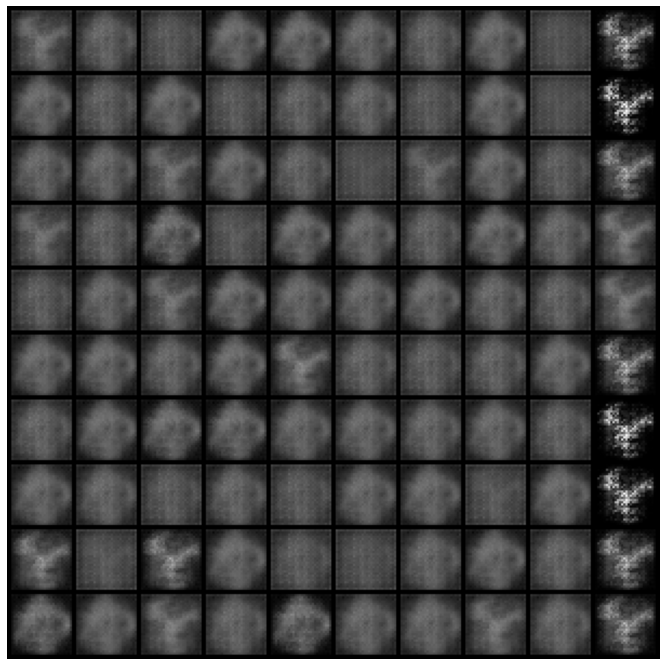}
	\end{subfigure}

	\begin{subfigure}{0.95\textwidth}
		\centering
		\includegraphics[height=1.2in]{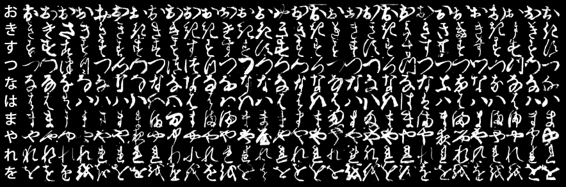}
	\end{subfigure}
	\caption{(a) Original KMNIST characters and CausalMIL reconstructions. The original digits are randomly sampled from the test set. 
		The figure in the last row contains original Kuzushiji characters for all 10 classes: the first column of this subfigure shows each character in its modern hiragana form, while the other columns are each character's handwritten forms.
		These figures are best viewed when zoomed.}
	\label{MNIST-details}
\end{figure*}

\subsection{Ablation Results}
\begin{figure*}[!t]
	\centering
	\begin{subfigure}{0.24\textwidth}
		\includegraphics[height=1.2in]{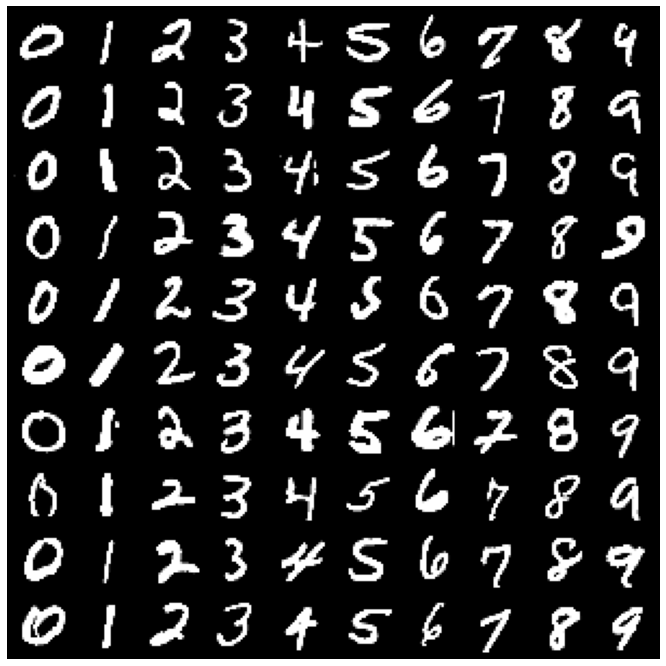}
		\caption{AE Original.}
	\end{subfigure}
	\begin{subfigure}{0.24\textwidth}
		\includegraphics[height=1.2in]{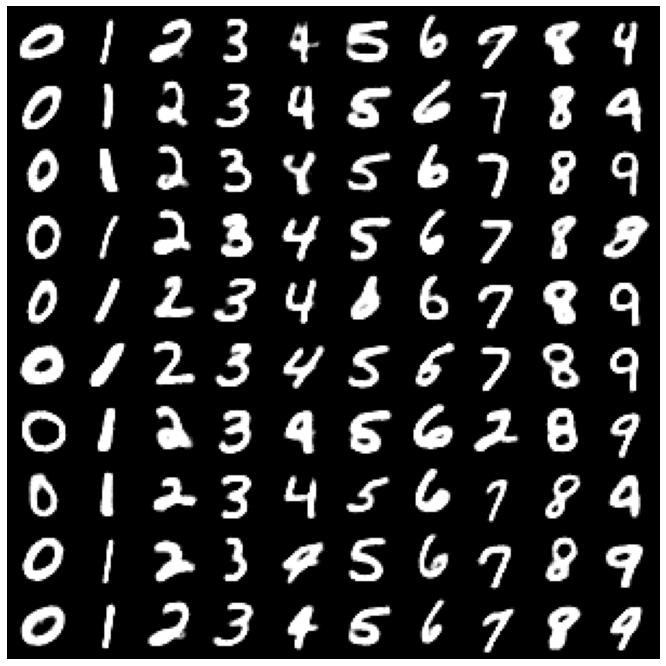}
		\caption{AE Recon.}
	\end{subfigure}
	\begin{subfigure}{0.24\textwidth}
		\includegraphics[height=1.2in]{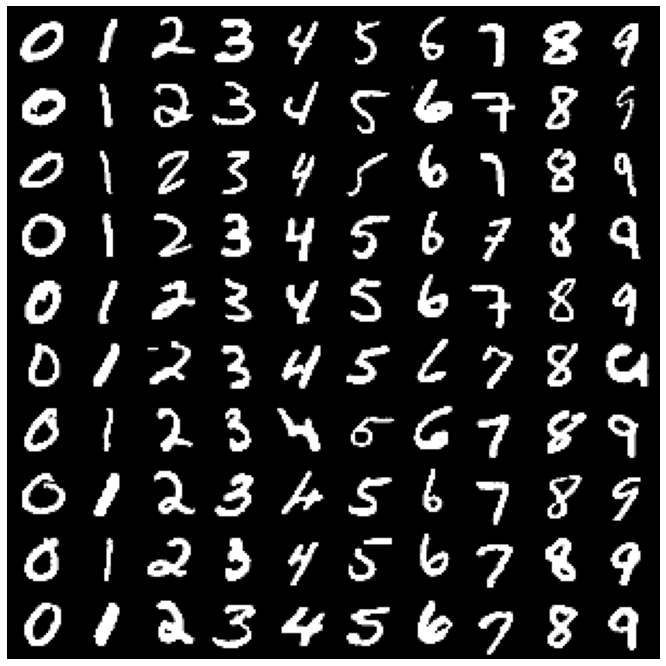}
		\caption{miAE Original.}
	\end{subfigure}
	\begin{subfigure}{0.24\textwidth}
		\includegraphics[height=1.2in]{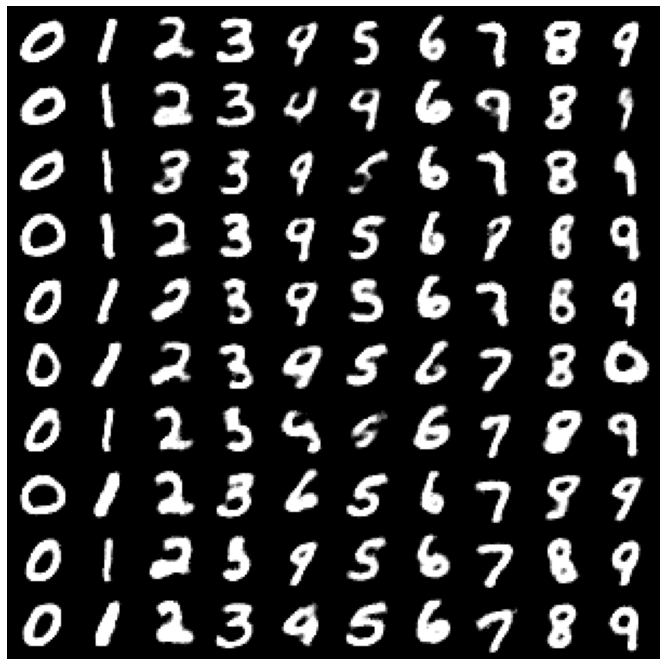}
		\caption{miAE Recon.}
	\end{subfigure}
	
	\begin{subfigure}{0.24\textwidth}
		\includegraphics[height=1.2in]{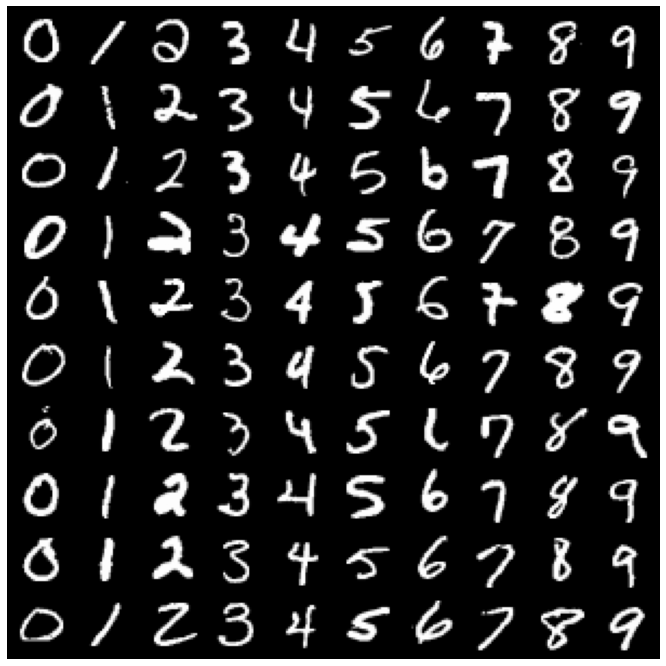}
		\caption{miVAE Original.}
	\end{subfigure}
	\begin{subfigure}{0.24\textwidth}
		\includegraphics[height=1.2in]{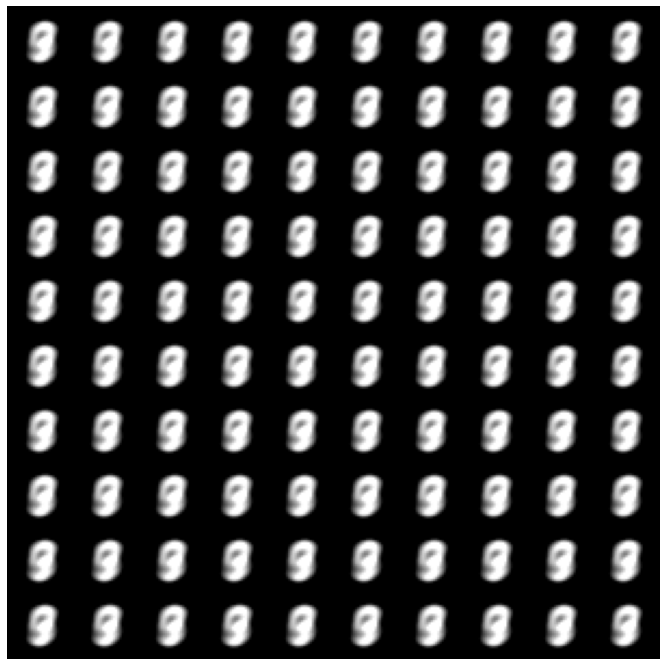}				
		\caption{miVAE Recon.}
	\end{subfigure}
	\begin{subfigure}{0.24\textwidth}
		\includegraphics[height=1.2in]{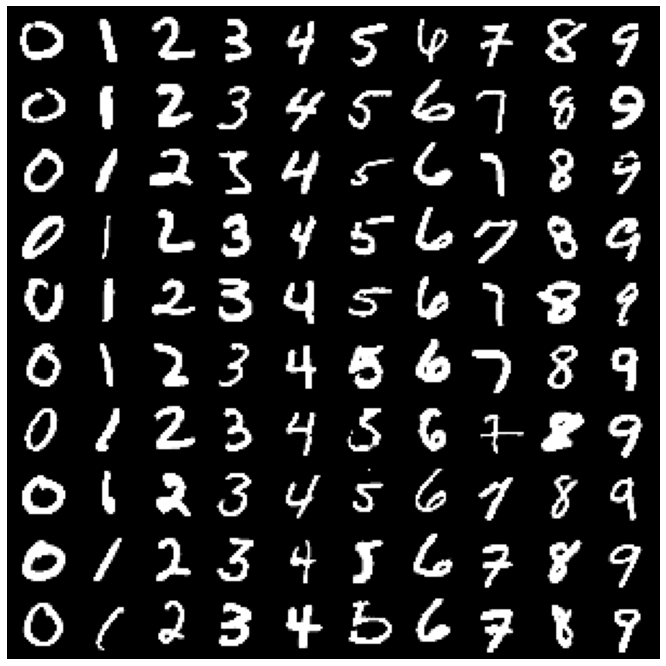}
		\caption{MIVAE Original.}
	\end{subfigure}
	\begin{subfigure}{0.24\textwidth}
		\includegraphics[height=1.2in]{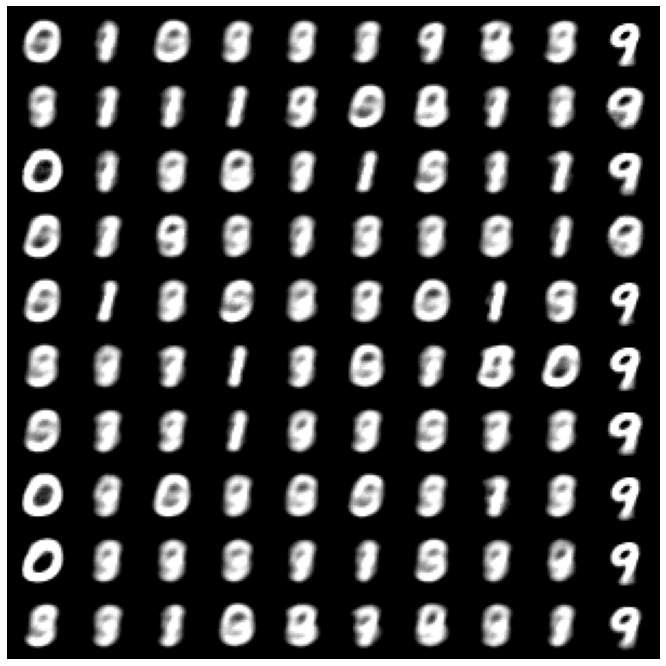}				
		\caption{MIVAE Recon.}
	\end{subfigure}
	
	\begin{subfigure}{0.24\textwidth}
		\includegraphics[height=1.2in]{figures/MNIST_original_9.png}
		\caption{CausalMIL Original.}
	\end{subfigure}
	\begin{subfigure}{0.24\textwidth}
		\includegraphics[height=1.2in]{figures/MNIST_recon_9.png}
		\caption{CausalMIL Recon.}
	\end{subfigure}

	\caption{(a) Original digits and reconstructions learned for the digit `9' of ablations and CausalMIL.
		These figures are best viewed when zoomed.}
	\label{ablation}
\end{figure*}

In this section we conduct ablation studies of CausalMIL. Specifically, in Figure \ref{ablation} we show (1) the reconstructions of a standard Autoencoder (AE); 
(2) the reconstructions of CausalMIL without the KL regularization term (miAE, a multi-instance Autoencoder with max-pooling);
(3) the reconstructions of a non-targeted version of CausalMIL (miVAE  a multi-instance Variational Autoencoder and we denote it as miVAE); and 
(4) the reconstructions of MIVAE \cite{Zhang2021} (another multi-instance Variational Autoencoder that learns an instance-specific latent factor and a bag-level shared latent factor similar to group-based VAE methods). 

From the reconstruction results of AE, miAE and miVAE, we can see that both targeted reconstruction and the bag-dependent KL divergence is necessary for learning meaningful representations: without the bag conditional KL term, the representations learned by miAE is similar to a vanilla Autoencoder.
Without the targeted reconstruction, the latent representation learned by miVAE is indistinguishable to human eyes. 

From the reconstructions of MIVAE (which is trained on bags with `9' versus those without), we can see that it does learn the representation of `9'.
However, the representations are very noisy: firstly, it also learns the representations of other digits, such as `0' and `1', and many other digits have been incorrectly reconstructed to `0' and `1'. Secondly, it reconstructs digits that do not resemble '9', such as the `6' at the 1st row into `9', 7th column. 
Theses noises in the latents are caused by the fact that MIVAE does not have identifiability guarantee. 
%performs worse than CausalMIL in the quantitative results. 

%\subsection{Sensibility to Latent Dimensions}
%In this section we study how different number of latent dimensions affect the latent representation learned by CausalMIL. Specifically, we present the reconstructions of FashionMNIST-bags using a latent dimensions of $2,4,8,16,32$ and $64$.

\subsection{Sensitivity to Bag Sizes and Witness Rates}
\begin{figure}[!t]
	\centering
	\begin{subfigure}{0.48\linewidth}
		\includegraphics[height=1.2in]{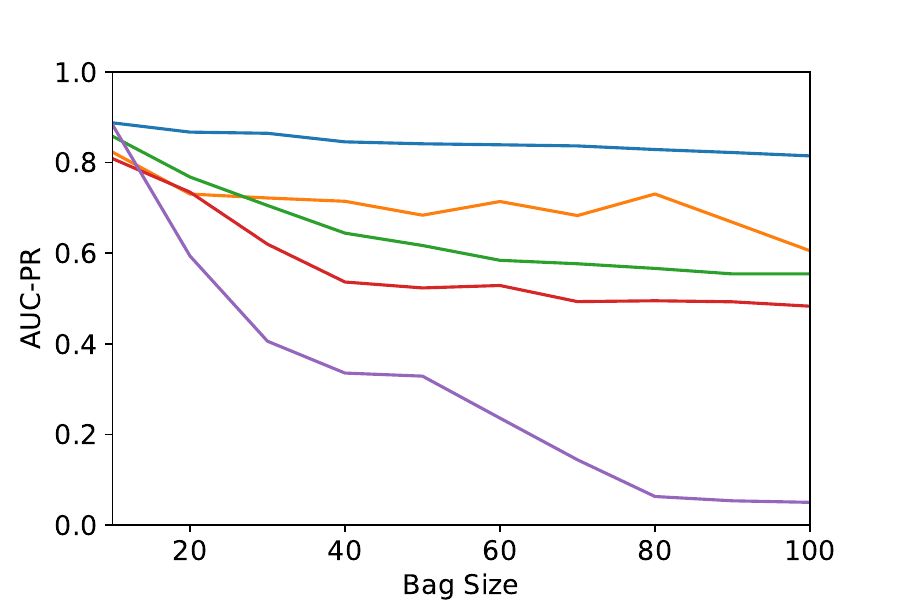}
			\caption{Bag size.}
	\end{subfigure}
	\begin{subfigure}{0.48\linewidth}
		\includegraphics[height=1.2in]{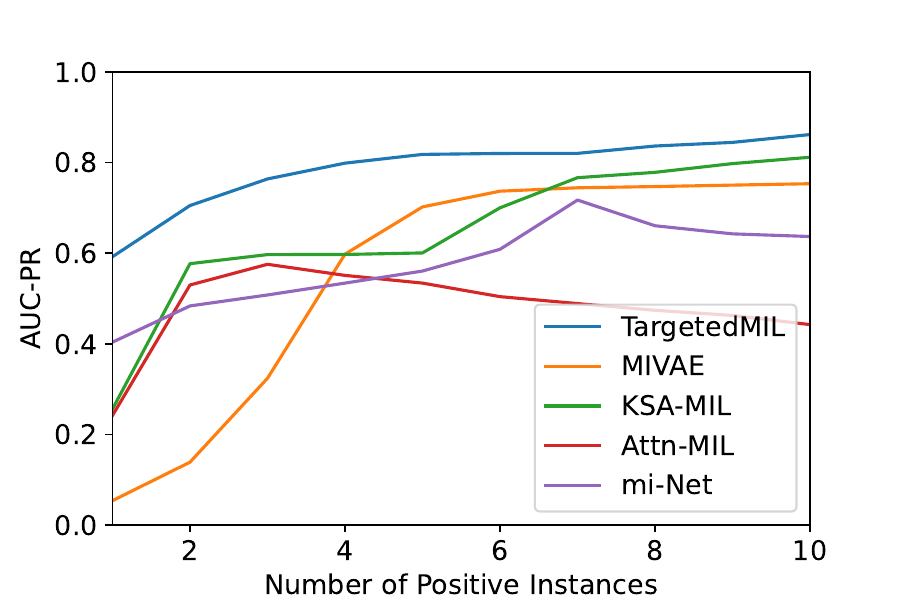}
			\caption{Witness rate.}
	\end{subfigure}
	\caption{ The average AUC-PR of different bag sizes (left) and witness rates (right) on ten FashionMNIST-bags.}
	\label{bag_size}
\end{figure}

Figure \ref{bag_size} shows how different bag sizes and witness rates (number of positive instances per bag) affect the instance prediction performances of different MIL algorithms. 
From the bag size results, we can see that the performances of all compared MIL algorithms decrease as the bag size increases, which is not surprising as it affects the levels of inexactness in the supervision; 
however, the performances drop of CausalMIL is significantly less severe than the compared ones. 
From the witness rate results, it is interesting to observe that larger witness sometimes decreases performances for algorithms that consider instances as independent (e.g., mi-Net and Attn-MIL). 
Again, CausalMIL performs consistently better than the compared algorithms.

\section{Proof of Identifiability}
In this section, we provide the proof for the latent identifiability of the model in Figure \ref{model}. 
Our proof consists of three main components. It is worth noting that compared to the proof of iVAE \cite{Khemakhem2020}, the important differences in our proof are located in the third step.

In the first step, we use the first assumption in Theorem \ref{thm-identifiability} to demonstrate that the observed data distributions are equivalent to the noiseless distributions. 
Specifically, suppose that we have two sets of parameters $(\bm{f}, \bm{T}, \bm{\lambda})$ and $(
\tilde{\bm{f}}, \tilde{\bm{T}} , \tilde{\bm{\lambda}})$, with a change of variable $\overline{\bm{x}} = \bm{f}(\bm{z}) = \tilde{\bm{f}}(\bm{z})$, we have:
\begin{equation}
	\tilde{p}_{\bm{T},\bm{\lambda}, \bm{f}, \bm{B}, y}(\bm{x}) = \tilde{p}_{ \tilde{\bm{T}}, \tilde{\bm{f}}, \tilde{\bm{\lambda}}, \bm{B}, y}(\bm{x}).
\end{equation}
\begin{proof}
	For simplicity of notations we denote $(\bm{B},y)$ by $\bm{U}$, we have
	\begin{align}
		p_{\bm{\theta}} (\bm{x} \vert \bm{U}) & = p_{\bm{\tilde{\theta}}} (\bm{x} \vert \bm{U}) \\
		\implies \int p_\epsilon(\bm{x} -\bm{f} (\bm{z}))p_{\bm{T},\bm{\lambda}}(\bm{z}\vert\bm{U})d\bm{z} 
		& = \int p_\epsilon(\bm{x} -\bm{\tilde{f}} (\bm{z}))p_{\bm{\tilde{T}},\bm{\tilde{\lambda}}}(\bm{z}\vert\bm{U})d\bm{z} \\
		\implies \int p_\epsilon(\bm{x} -\bm{\bar{x}}) p_{\bm{T},\bm{\lambda}}( \bm{f}^{-1} (\bm{\bar{x}}\vert \bm{U}) vol(J_{f^{-1}}(\bm{\bar{x}})) d\bm{\bar{x}} 
		& = \int p_\epsilon(\bm{x} -\bm{\bar{x}}) p_{\bm{T},\bm{\lambda}}( \bm{\tilde{f}}^{-1} (\bm{\bar{x}}\vert \bm{U}) vol(J_{\tilde{f}^{-1}}(\bm{\bar{x}})) d\bm{\tilde{x}} 
		\label{eq_jacobian}\\
		\implies \int p_\epsilon(\bm{x} -\bm{\bar{x}}) \tilde{p}_{\bm{T},\bm{\lambda}, \bm{f}, \bm{U}}(\bm{\bar{x}}) d\bm{\bar{x}} & = \implies \int p_\epsilon(\bm{x} -\bm{\bar{x}}) \tilde{p}_{ \tilde{\bm{T}}, \tilde{\bm{f}}, \tilde{\bm{\lambda}}, \bm{U}}(\bm{\bar{x}}) d\bm{\bar{x}} '
		\label{eq_jacobian2}\\
		\implies  (\tilde{p}_{\bm{T},\bm{\lambda}, \bm{f}, \bm{U}} * p_\varepsilon) (\bm{x}) 
		&= (\tilde{p}_{ \tilde{\bm{T}}, \tilde{\bm{f}}, \tilde{\bm{\lambda}}, \bm{U}} * p_\varepsilon) (\bm{x}) 
		\label{eq_convolution}\\
		\implies F[\tilde{p}_{\bm{T},\bm{\lambda}, \bm{f}, \bm{U}}] (\bm{\omega}) \phi_\varepsilon(\bm{\omega}) & = F[\tilde{p}_{\tilde{\bm{T}}, \tilde{\bm{f}}, \tilde{\bm{\lambda}}}] (\bm{\omega}) \phi_\varepsilon(\bm{\omega}) 
		\label{eq_fourier}\\
		\implies  F[\tilde{p}_{\bm{T},\bm{\lambda}, \bm{f}, \bm{U}}] (\bm{\omega})  & = F[\tilde{p}_{\tilde{\bm{T}}, \tilde{\bm{f}}, \tilde{\bm{\lambda}}, \bm{U}}] (\bm{\omega})  \label{eq_assumtion1}\\
		\implies \tilde{p}_{\bm{T},\bm{\lambda}, \bm{f}, \bm{U}}(\bm{x}) &= \tilde{p}_{ \tilde{\bm{T}}, \tilde{\bm{f}}, \tilde{\bm{\lambda}}, \bm{U}} (\bm{x}). \label{eq_step1}
	\end{align}
where 
\begin{itemize}
	\item in Equation \ref{eq_jacobian}, $J$ denotes the Jacobian, $vol(B)=\sqrt{\det(B^T B)}$.
	\item in Equation \ref{eq_jacobian2}, we introduced
	\begin{equation}
	\tilde{p}_{\bm{T},\bm{\lambda}, \bm{f}, \bm{U}} \triangleq p_{\bm{T},\bm{\lambda}}(\bm{f}^{-1})(\bm{x} \vert \bm{U}) vol (J_{f^{-1}} (\bm{x})) \mathbb{I}(\bm{x})
	\label{eq_definition}
	\end{equation}
	\item in Equation \ref{eq_convolution}, $*$ denotes the convolution operator.
	\item in Equation \ref{eq_fourier}, $F$ denotes the Fourier transformation and $\phi_\varepsilon = F[p_\varepsilon]$. 
	\item in Equation \ref{eq_assumtion1}, $\phi_\varepsilon(\bm{w})$ is dropped because it is non-zero almost everywhere according to the first assumption of Theorem \ref{identifiability}.
\end{itemize}
\end{proof}

In the second step, the fourth assumption in Theorem \ref{identifiability} is used for removing all the terms that are a function of $\bm{x}$ or $\bm{B}$. By substituting $p_{\bm{T},\bm{\lambda}}$ with its exponential conditionally factorial form, we can show that:
\begin{equation}
	\bm{T} (\bm{f}^{-1}) (\bm{x}) = 	\bm{A} \bm{\tilde{T}} (\bm{\tilde{f}}^{-1}) (\bm{x}) + \bm{c}
\end{equation}
\begin{proof}
	Using Equation \ref{eq_definition} to substitute Eauation \ref{eq_step1}, we have
	\begin{equation}
	p_{\bm{T},\bm{\lambda}}(\bm{f}^{-1})(\bm{x} \vert \bm{U}) vol (J_{f^{-1}} (\bm{x})) \mathbb{I}(\bm{x}) = p_{\bm{\tilde{T}}},\bm{\tilde{\lambda}}(\bm{\tilde{f}}^{-1})(\bm{x} \vert \bm{U}) vol (J_{\tilde{f}^{-1}} (\bm{x})) \mathbb{I}(\bm{x}).
	\end{equation}
	Then, we can apply logarithm on the above equation and substitute $p_{\bm{T},\bm{\lambda}}$ with its definition in Equation \ref{eq_generative}, and obtain
	\begin{align}
		\log vol(J_{f^{-1}} (\bm{x})) \log Q(\bm{f}^{-1} \bm{x}) &- \log Z(\bm{U}) + \langle \bm{T}(\bm{f}^{-1}(\bm{x})) , \bm{\lambda}(\bm{U}) \rangle \\
		&= \log vol(J_{\tilde{f}^{-1}} (\bm{x})) \log \tilde{Q}(\bm{\tilde{f}}^{-1} \bm{x}) - \log \tilde{Z}(\bm{U}) + \langle \bm{\tilde{T}}(\bm{\tilde{f}}^{-1}(\bm{x})) , \bm{\tilde{\lambda}}(\bm{U}) \rangle 
	\end{align}
Let $\bm{U}^0, \cdots, \bm{U}^{k}$ be the $k+1$ points defined in the fourth assumption of Theorem \ref{thm-identifiability}, we can obtain $k+1$ equation. By subtracting the first equation from the remaining $k$ equations, we then obtain:
\begin{align}
	\langle \bm{T}(\bm{f}^{-1}(\bm{x})) , \bm{\lambda}(\bm{U}^l) &-\bm{\lambda}(\bm{U}^0)  \rangle + \log\frac{Z(\bm{U}^0)}{Z(\bm{U}^l)} \nonumber\\
	&= \langle \bm{\tilde{T}}(\bm{\tilde{f}}^{-1}(\bm{x})) , \bm{\tilde{\lambda}}(\bm{U}^l) -\bm{\tilde{\lambda}}(\bm{U}^0)  \rangle + \log\frac{\tilde{Z}(\bm{U}^0)}{\tilde{Z}(\bm{U}^l)},
\end{align}
where $l = 1, \cdots, k$. Let $\bm{b}\in \mathbb{R}^k$ in which $b_l = \log\frac{\tilde{Z}(\bm{U}^0) Z(\bm{U}^l)}{\tilde{Z}(\bm{U}^l) Z(\bm{U}^0)}$, we have
\begin{equation}
	L^T\bm{T}(\bm{f}^{-1}(\bm{x})) = \tilde{L} \bm{\tilde{T}} (\bm{\tilde{f}}^{-1}(\bm{x})) + \bm{b}.
\end{equation}
Finally, we multiply both side by $L^{-T}$ and obtain 
\begin{equation}
	\bm{T}(\bm{f}^{-1}(\bm{x})) = A \bm{\tilde{T}} (\bm{\tilde{f}}^{-1} (\bm{x})) + \bm{c},
	\label{eq_linear_transformation}
\end{equation}
where $A=L^{-T}L$ and $\bm{c}= L^{-T}\bm{b}$.
\end{proof}
In the third step, we show that the transformation $A$ is invertible, such that:
\begin{equation}
	(\bm{f}, \bm{T}, \bm{\lambda}) \sim (\tilde{\bm{f}}, \tilde{\bm{T}}, \tilde{\bm{\lambda}}),
\end{equation}
\begin{proof}
	We start by evaluating Equation \ref{eq_linear_transformation} at $k+1$ points of $\bm{z}_l, \bm{x}_l$ and obtain $k+1$ equations. Then, we subtract the first equation from the remaining $k+1$ equations:
	\begin{align}
		[\bm{T}(\bm{z}_1) -\bm{T}(\bm{z}_0), &\cdots, \bm{T}(\bm{z}_k) -\bm{T}(\bm{z}_0) ] \nonumber\\
		=A[\bm{\tilde{T}}(\bm{\tilde{f}}^{-1} (\bm{x}_1)) - \bm{\tilde{T}}(\bm{\tilde{f}}^{-1} (\bm{x}_0)) , \cdots, \bm{\tilde{T}}(\bm{\tilde{f}}^{-1} (\bm{x}_l)) - \bm{\tilde{T}}(\bm{\tilde{f}}^{-1} (\bm{x}_0))   ].
	\end{align}
Next we only need to show that for $\bm{z}_0$ there exist $k$ points $\bm{z}_1, \cdots,\bm{z}_k$ such that the columns are linear independent, which can be proven by contradiction. Suppose that there exists no such $\bm{z}_1,\cdots,\bm{z}_k$, then $\langle \bm{T}(\bm{z}) - \bm{T}(\bm{z}_0) ,\bm{\lambda} \rangle = 0$ and thus $\bm{T}(\bm{z}) = \bm{T}(\bm{z}_0) = \text{const}$. This contradicts with the assumption that the prior distribution is strongly exponential. Therefore, there must exist $k+1$ points such that the transformation is invertible. 
\end{proof}

\section{Implementation and Experiment Details}
\subsection{Details and Parameters}
CausalMIL is implemented using PyTorch 1.12. 
In the implementation, the term $\log p_{\bm{\vartheta}}(\bm{B}\vert\bm{z})$ is omitted from the ELBO as $\bm{z}^c$ is independent of $\bm{B}$. 
The unconditional priors are assumed to be i.i.d sampled from $\mathcal{N}(\bm{0},\bm{1})$. 
Furthermore, we enforce the bag transformation mapping $\phi$ to share the same parameters as those of the encoder. 

For all experiments, the CausalMIL models are trained with AdamW optimizer for 200 epochs, and the best epochs are selected simply according to the training ELBO. The parameters and the best epoch of CausalMIL are also tuned using the training ELBO.
To show that our results are indeed identifiable instead of cherry picking the reconstructions, for all of the MNIST-bags, FashionMNIST-bags, KuzushijiMNIST-bags experiments the parameters are fixed to the same value tuned according to MNIST-bags, i.e., we set learning rate to $1e-3$, weight for max-pooling to $1000$, weight for the KL divergences to $1$. The latent dimensions for $\bm{z}$ is set to $24$ for all datasets.
The optimization methods and parameters for the compared methods selected according to the strategy provided in their publicly available implementations and in their publications. We also performed extensive grid search for parameters of the compared algorithms. 
however, we find that this does not significantly improve their performances.

%We would also like to highlight that the parameters for CausalMIL is only selected according to one task. Specifically, its parameters are selected on MNIST-bags with `1' as the target concept and the others digits as negative concepts. The selected parameters are then used across all other tasks in MNIST-bags, FashionMNIST-bags, and KMNIST-bags. These results demonstrate that the causal representations and performances of CausalMIL is due to its model design and is not sensitive to parameter choices.

\subsection{Network structures}
For the MNIST, FashionMNIST, and KuzushijiMNIST bags, we use the convolutional encoder and decoder illustrated in Table \ref{conv}.
\begin{table}[h]
	\centering
	\caption{Encoding and decoding network structures for MNIST, FashionMNIST and KuzushijiMNIST bags.}
	\begin{tabular}{|c|c|}
		\hline
		Layer & Type \\
		\hline
		1 & conv2d(4,2,1)-32 + BatchNorm + ReLU\\
		\hline
		2 & conv2d(4,2,1)-128 + BatchNorm + ReLU\\
		\hline
		3 & conv2d(7,1,0)-512 + BatchNorm + ReLU\\
		\hline
	\end{tabular}
	
	\begin{tabular}{|c|c|}
		\hline
		Layer & Type \\
		\hline
		1 & ConvTranspose2d(7,1,0)-512 + BatchNorm +ReLU\\
		\hline
		2 & ConvTranspose2d(4,2,1)-128 + BatchNorm + ReLU\\
		\hline
		3 & ConvTranspose2d(4,2,1)-32 + BatchNorm  + ReLU\\
		\hline
	\end{tabular}
	\label{conv}
\end{table}

%\subsubsection{Colon Cancer Histopathology Images}
%\subsubsection{Histopathology Images}
%Now we examine the instance prediction performance of CausalMIL with a real-world medical diagnosis dataset.
%Histopathology with hematoxylin and eosin stained whole-slide images is an important MIL application since supervised learning algorithms require pixel-level annotations that are difficult to obtain. 
%Therefore, a label efficient multi-instance approach using only image-level labels is essential. 
%We evaluate CausalMIL against state-of-the-art deep learning based MIL algorithms for predicting instance label using bag supervised images from the colon cancer dataset provided in \cite{Sirinukunwattana2016}.
%
%This dataset contains 100 images where each one is obtained from a patient's tissue of either normal or malignant regions. 
%For each image bag, the instances are generated as patches $27\times 27$ pixels using markings of major nuclei for each cell. A total amount of 22,444 instances (~220 instances per bag) are provided with ground truth instance class labels, i.e. epithelial or not. Epithelial is an important cell development stage of invasive carcinoma.
%A bag is considered to be positive if it contains at least one patch that is epithelial.

For the Colon Cancer histopathology dataset, we use the same convoluntional network structures as the compared methods for fairness of comparison. For encoding the images, we used the following convolutional network (Table \ref{conv_colon} (left)) as used in \cite{Ilse2018}. For decoding the from the latent factors, we use a corresponding de-convolutional network as specified in Table \ref{conv_colon} (right). 
%The parameters grids for image datasets are provided in Table \ref{parameters_image}.

\begin{table}[h]
	\centering
	\caption{Network structures for the Colon Cancer dataset.}
	\begin{tabular}{|c|c|}
		\hline
		Layer & Type \\
		\hline
		1 & conv(4,1,0)-36 + ReLU\\
		\hline
		2 & maxpool(2,2) \\
		\hline
		3 & conv(3,1,0)-48 + ReLU\\
		\hline
		4 & maxpool(2,2) \\
		\hline
	\end{tabular}
	~
	\begin{tabular}{|c|c|}
		\hline
		Layer & Type \\
		\hline
		1 & Upsampling(10)\\
		\hline
		2 & ConvTranspose(3,1,0)-48 +ReLU\\
		\hline
		3 & Upsampling(5)\\
		\hline
		4 &   ConvTranspose(4,1,0)-36 +ReLU \\
		\hline
	\end{tabular}
	\label{conv_colon}
\end{table}

Furthermore, for the biased datasets we use the same network structures as in the causal invariant representation learning literature \cite{Arjovsky2019,Ahuja2020,lu2022} as illustrated in Table \ref{conv_biased}.

\begin{table}[h]
	\centering
	\caption{Encoding and decoding network structures for the distribution biased datasets \cite{Arjovsky2019}.}
	\begin{tabular}{|c|c|}
		\hline
		Layer & Type \\
		\hline
		1 & conv2d(3,2,1)-32  + ReLU\\
		\hline
		2 & conv2d(3,2,1)-32  + ReLU\\
		\hline
		3 & conv2d(3,1,0)-32  + ReLU\\
		\hline
	\end{tabular}
	
	\begin{tabular}{|c|c|}
		\hline
		Layer & Type \\
		\hline
		1 & ConvTranspose2d(3,2, 1)-32  +ReLU\\
		\hline
		2 & ConvTranspose2d(3,2,1)-32  + ReLU\\
		\hline
		3 & ConvTranspose2d(3,2,1)-3   + ReLU\\
		\hline
	\end{tabular}
	\label{conv_biased}
\end{table}

\subsection{Data and Code Availability}
Pre-processed multi-instance bags for the Colon Cancer dataset can also be downloaded directly from \url{https://drive.google.com/file/d/1RcNlwg0TwaZoaFO0uMXHFtAo_DCVPE6z/view?usp=sharing}. 
The original images of this dataset are available at \url{https://warwick.ac.uk/fac/sci/dcs/research/tia/data/crchistolabelednucleihe/}.

%The code for reproducing the experiments can be found . 
For the compared methods, their implementations can be found at: \\
mi-Net is implemented at \url{https://github.com/yanyongluan/MINNs} \\
AttentionMIL is implemented at \url{https://github.com/AMLab-Amsterdam/AttentionDeepMIL}\\
KernelSelfAttention-MIL is implemented at \url{https://github.com/gmum/Kernel_SA-AbMILP}\\
MIVAE is implemented at \url{https://github.com/WeijiaZhang24/MIVAE}.

\end{document}